\documentclass{llncs}

\usepackage[dvips]{graphics}
\usepackage{epsfig}
\usepackage{amsfonts}
\usepackage{amssymb}
\usepackage{amsmath}
\usepackage{color}
\usepackage{url}

\usepackage{algorithmic,algorithm}

\newcommand{\commentaire}[1]{ }

\newcommand{\Real}{\mathop{\rm I\kern-.2emR}}
\newcommand{\Nat}{\mathop{\rm I\kern-.2emN}}
\renewcommand{\P}{\mathop{\rm I\kern-.2emP}}

\begin{document}

\mainmatter

\title{Analyzing the Effect of Objective Correlation on the Efficient Set of MNK-Landscapes}

\titlerunning{On the Efficient Set of MNK-Landscapes with Objective Correlation}

\author{S\'ebastien Verel\inst{1,3} \and Arnaud Liefooghe\inst{2,3} \and \\ Laetitia Jourdan\inst{3} \and Clarisse~Dhaenens\inst{2,3}}

\authorrunning{S. Verel, A. Liefooghe, L. Jourdan and C. Dhaenens}   

\institute{University of Nice Sophia Antipolis -- CNRS, France 
\and Universit\'e Lille 1, LIFL -- CNRS, France 
\and INRIA Lille-Nord Europe, France\\
\email{verel@i3s.unice.fr, arnaud.liefooghe@univ-lille1.fr, laetitia.jourdan@inria.fr, clarisse.dhaenens@lifl.fr}
}

\maketitle

\begin{abstract}
In multiobjective combinatorial optimization,
there exists two main classes of metaheuristics, based either on multiple aggregations, or on a dominance relation.
As in the single-objective case, 
the structure of the search space can explain the difficulty for multiobjective metaheuristics, and guide the design of such methods.
In this work we analyze the properties of multiobjective combinatorial search spaces.
In particular, we focus on the features related the efficient set, and we pay a particular attention to the correlation between objectives.
Few benchmark takes such objective correlation into account.
Here, we define a general method to design multiobjective problems with correlation. 
As an example, we extend the well-known multiobjective $NK$-landscapes. 
By measuring different properties of the search space, 
we show the importance of considering the objective correlation on the design of metaheuristics.
\end{abstract}

\section{Introduction}
\label{sec:intro}

Multiobjective combinatorial optimization (MoCO) problems, where several criteria have to be optimized simultaneously,
receive more and more interest in the field of search algorithms.
One of the main issues in multiobjective optimization is the Pareto dominance relation, which gives a partial order between feasible solutions.
Roughly speaking, a given solution dominates another solution if it is better according to all objective functions.
A possible approach in solving a multiobjective problem consists in finding the whole set of non-dominated solutions,
called the \emph{efficient set}, or a subset that is close to it.
This efficient set plays a central role in the structure of the search space.

The design of metaheuristics for multiobjective combinatorial optimization is a real challenge, as it is problem-dependent.
Like in single-objective optimization, the structure of the search space can explain the ability of multiobjective metaheuristics.
Two main classes of multiobjective metaheuristics can be distinguished.
The first ones, known as scalar approaches, are based on multiple scalarized aggregations of the objective functions.
However, they are only able to find a subset of efficient solutions, called supported efficient solutions.
The second ones, known as Pareto-based approaches, directly or indirectly focus the search on the Pareto dominance relation.
Moreover, when the size of the efficient set is too large, 
a metaheuristic should manipulate a limited-size solution set during the search, and this limit is related to the size of the efficient set.
In addition, connectedness is related to the property that efficient solutions are connected with respect to a neighborhood relation
\cite{ehrgott1997}.
When connectedness holds, it becomes possible to find the whole efficient set 
by iteratively exploring the neighborhood of the current approximation, initialized with at least one efficient solution.
This strategy is often used explicitly, or implicitly by Pareto-based approaches.
For the design of metaheuristics for MoCO, three main questions, related to the efficient set properties, are of our interest in this paper:
\begin{itemize}
\item[($i$)] What is the cardinality of the efficient set?
Can we pretend to identify or approximate the whole set of efficient solutions, or should we consider a mechanism to bound the size of the approximation set?
\item[($ii$)] How many efficient solutions are supported?
Is a scalar approach able to identify or approximate enough efficient solutions?
\item[($iii$)] Are efficient solutions connected with respect to a neighborhood operator?
Is it possible to identify or approximate additional efficient solutions by a simple local search initialized with a subpart of the efficient set?
\end{itemize}
In particular we want to study such properties according to the objective correlation,
as it seems to largely affect the solutions of MoCO problems~\cite{Mote1991} and the behavior of metaheuristics~\cite{paquete2006}.
Few benchmark takes the correlation between objectives into account. 
To the best of our knowledge, the multiobjective quadratic assignment problem~\cite{knowles2003} should be the single one.
In this problem, a parameter can tune the correlation between different pairs of objectives.
Another well-known benchmark, the multiobjective $NK$-landscapes~\cite{aguirre2007} facilitate the study of problem structure in multiobjective optimization. 
In this class, the epistatic degree, which is the degree of non-linearity of the problem, can be tuned very precisely.
In this work, in order to study the problem structure, and in particular the structure of the efficient set, 
we define a general method to tune the correlation between all pairs of objectives very precisely. 
As an example, we define the multiobjective $\rho MNK$-landscapes, an extension of multiobjective $NK$-landscapes with objective correlation.
With such a benchmark,  
we can study the problem structure according to the objective space dimension, the epistasis and especially the objective correlation, 
and then highlight some guidelines for the design of efficient multiobjective metaheuristics.

In summary, the contributions of this work can be stated as follows.
First, we propose a method to precisely tune the correlation between objective functions.
It is applied to the design of $MNK$-landscapes, but it can easily be generalized to other problems.
Second, we show the influence of the objective correlation on some properties of the efficient set (and its image in the objective space):
its size, the proportion of supported solutions, and the connectedness of efficient solutions.
Third, we bring those properties with the design of local search metaheuristics
in order to help the practitioner to make proper choices between several classes of methodologies.
The reminder of the paper is organized as follows. 
Section 2 is dedicated to multiobjective combinatorial optimization, multiobjective metaheuristics,
as well as single- and multi-objective $NK$-landscapes.
Section 3 presents the design of $\rho MNK$-landscapes.
We conduct a theoretical analysis and an experimental study to show the sharpness of the objective correlation.
Section 4 deeply analyzes the efficient set structure on this new class of problems 
according to the objective space dimension, the non-linearity and especially the objective correlation.
The consequence on the design of multiobjective metaheuristics are discussed in the last section.

\section{Background}

\subsection{Multiobjective Combinatorial Optimization}
A large number of real-world optimization problems are multiobjective by nature, because several criteria have to be considered simultaneously. 
A MoCO problem can be defined by a set of $M \geq 2$ objective functions $(f_1, f_2,\dots, f_M)$,
and a discrete set $X$ of feasible solutions in the \emph{decision space}.
Let $Z  =  f(X) \subseteq \Real^M$ be the set of feasible outcome vectors in the  \emph{objective space}.  
In a maximization context, a solution $x \in X$ dominates a solution $x' \in X$, denoted by $x \succ x'$,
iff $\forall i \in \{1,2,\dots,M\}$, $f_i(x) \geq f_i(x')$ and $\exists j \in \{1,2,\dots,M\} $ such as $f_j(x) > f_j(x')$.
A solution $x \in X$ is said to be \emph{efficient} (or \emph{non-dominated}, \emph{Pareto optimal}), 
if there does not exist any other solution $x^{'} \in X$ such that $x^{'}$ dominates $x$.
The set of all efficient solutions is called the \emph{efficient set} (or \emph{Pareto optimal set}), denoted by $X_E$,
and its mapping in the objective space is called the \emph{Pareto front}.
A possible approach in MoCO is to identify a minimal complete efficient set,
{\itshape i.e.} one efficient solution mapping to each point of the Pareto front.

However, generating the entire efficient set of a MoCO problem is often infeasible for two main reasons~\cite{ehrgott2005}.
First, for most MoCO problems, the number of efficient solutions is known to be exponential in the size of the problem instance.
In that sense, most MoCO problems are said to be \emph{intractable}.
Second, deciding if a feasible solution belongs to the efficient set is NP-complete for numerous MoCO problems,
even if none of its single-objective counterpart is NP-hard.
Therefore, the overall goal is often to identify a good efficient set approximation.
To this end, metaheuristics in general, and evolutionary algorithms in particular, have received a growing interest since the late eighties,
and multiobjective metaheuristics still constitute an active research area.

\subsection{Metaheuristics for Multiobjective Combinatorial Optimization}
\label{sec:moco}
Two main classes of metaheuristics for MoCO can be distinguished, see for instance \cite{paquete2007}.
The first ones, known as scalar approaches, are based on multiple scalarized aggregations of the objective functions.
The second ones, known as Pareto-based approaches, directly or indirectly focus the search on the Pareto dominance relation (or a slight modification of it).
These two kinds of approaches can also be hybridized in a two-phase way.

Initial approaches dealing with MoCO are based on successive transformations of the original multiobjective problem into single-objective ones
by means of a scalarization strategy.
Most of the time, \emph{scalar approaches} are based on a weighted-sum aggregation of the objective functions, that can be defined as follows.
$\forall x \in X$: $f_{\lambda}(x)=\sum_{i=1}^{M}{\lambda_i \ f_i(x)}$ where $\lambda_i > 0$ for all $i \in \{1, \dots, M\}$.
The problem is now to identify a (single) solution that maximizes $f_\lambda$.
For any given weighting coefficient vector $\lambda$, if $x^\star = arg\max_{x \in X} f_{\lambda}(x)$, then $x^\star$ is an efficient solution.
Multiple weighting coefficient vectors can be iteratively defined so that several non-dominated solutions are identified (or approximated). 
For each scalarization, the corresponding solution is incorporated into an approximation set,
whose dominated solutions are then discarded.
However, in the combinatorial case, a number of efficient solutions are not optimal for any definition of $f_\lambda$.
They are known as \emph{non-supported (efficient) solutions}.
On the contrary, there exists \emph{supported (efficient) solutions} whose corresponding objective vectors are located on the convex hull of the Pareto front.
The set of all supported efficient solutions will be denoted by $X_{SE}$.
As a consequence, the proportion of non-supported solutions over the efficient set has a direct implication on the ability of scalar approaches
to find a proper non-dominated set approximation.

Over the years, other types of approaches were proposed.
They are based on the explicit or implicit use of the Pareto dominance relation, that allows to define a partial order between feasible solutions.
The basic idea is to maintain a set solutions (typically a population or an archive of mutually non-dominated solutions).
The content of this set is then iteratively updated with new solutions built by means of variation or neighborhood operators. 
The update of this set is based on a specific decision on which solutions to accept or to choose for further manipulation.
This process is iterated until no further improvement is possible or another stopping condition is fulfilled.
In the end, this set corresponds to the approximation outputted by the algorithm.
The implicit goal is to identify an approximation whose image in the objective space is ($i$) close to and ($ii$)~well-spread along the Pareto front. 
However, as the number of efficient solutions is often intractable,
we generally have to design specific strategies to limit the size of the approximation set \cite{knowles2004}.
As a consequence, the \emph{cardinality} of the efficient set also plays a major role on the design of multiobjective metaheuristics.

More recently, the neighborhood structure of the efficient set has been claimed to play a crucial role for the development of efficient metaheuristics.
One of these properties is known as \emph{connectedness}~\cite{ehrgott1997,gorski2006}.
Let us define a graph such that each node represents an efficient solution,
and an edge connects a pair of nodes if the corresponding solutions are neighbors with respect to a given neighborhood operator
\cite{ehrgott1997}.
This graph is called the \emph{efficient graph}.
A neighborhood operator is a function $\mathcal{N}: X \rightarrow 2^X$ that assigns a set of solutions $\mathcal{N}(x) \subset X$ to any solution $x \in X$.
$\mathcal{N}(x)$ is called the \emph{neighborhood} of $x$, and a solution $x' \in \mathcal{N}(x)$ is called a \emph{neighbor} of $x$.
The efficient set is said to be \emph{connected} if there exists a path between every pair of nodes in the graph.
In other words, each efficient solution is located in the neighborhood of at least one other solution from the efficient set.
This property has later been extended to the notion of cluster
by introducing an arbitrary distance separating two efficient solutions \cite{paquete2009}.
When connectedness holds, it becomes possible to find all the efficient solutions by means of the iterative exploration of the neighborhood of the current approximation
by starting with one (or more) solution(s) from the efficient set.
This gives rise to a \emph{two-phase} approach:
($i$) identify a number of (typically supported) non-dominated solutions
($ii$) improve the set of non-dominated solutions by exploring their neighborhood.

\subsection{$NK$- and $MNK$-Landscapes}
The family of $NK$-landscapes \cite{kauffman93} is a problem-independent model used for constructing multimodal landscapes.
$N$ refers to the number of (binary) genes in the genotype ({\itshape i.e.} the string length)
and $K$ to the number of genes that influence a particular gene from the string (the epistatic interactions).
By increasing the value of $K$ from 0 to $(N-1)$, $NK$-landscapes can be gradually tuned from smooth to rugged.
The fitness function (to be maximized) of a $NK$-landscape $f_{NK}: \lbrace 0, 1 \rbrace^{N} \rightarrow [0,1)$ is defined on binary strings with $N$ bits. 
An `atom' with fixed epistasis level is represented by a fitness component $f_i: \lbrace 0, 1 \rbrace^{K+1} \rightarrow [0,1)$ associated to each bit $i \in N$.
Its value depends on the allele at bit $i$ and also on the alleles at $K$ other epistatic positions ($K$ must fall between $0$ and $N - 1$).
The fitness $f_{NK}(x)$ of a solution $x \in \lbrace 0, 1 \rbrace^{N}$ corresponds to the mean value of its $N$ fitness components $f_i$: \label{defNK}
$$ f_{NK}(x) = \frac{1}{N} \sum_{i=1}^{N} f_i(x_i, x_{i_1}, \ldots, x_{i_K})
$$ where $\lbrace i_1, \ldots, i_{K} \rbrace \subset \lbrace 1, \ldots, i-1, i+1, \ldots, N \rbrace$.
Several ways have been proposed to set the $K$ bits from the bit string of size $N$.
Two possibilities are mainly used: adjacent and random neighborhoods.
With an adjacent neighborhood, the $K$ bits nearest to the bit $i \in N$ are chosen (the genotype is taken to have periodic boundaries). With a random neighborhood,
the $K$ bits are chosen randomly on the bit string.
Each fitness component $f_i$ is specified by extension,
\textit{i.e.} a number $y^i_{x_i, x_{i_1}, \ldots, x_{i_K}}$ from $[0, 1)$ is associated with each element
$(x_i, x_{i_1}, \ldots, x_{i_K})$ from $\lbrace 0, 1 \rbrace^{K+1}$.
Those numbers are uniformly distributed in the range $[0, 1)$. 

More recently, a multiobjective variant of $NK$-landscapes (namely $MNK$- landscapes) \cite{aguirre2007}
have been defined with a set of $M$ fitness functions:
$$\forall m \in [1,M],\ f_{NK_{m}}(x) = \frac{1}{N} \sum_{i=1}^{N} f_{m, i}(x_i, x_{i_{m, 1}}, \ldots, x_{i_{m, K_{m}}})$$
The numbers of epistasis links $K_{m}$ can theoretically be different for each fitness function. 
But in practice, the same epistasis degree $K_{m}=K$ for all $m \in [1,M]$ is used.
Each fitness component $f_{m, i}$ is specified by extension
with the numbers $y^{m,i}_{x_i, x_{i_{m, 1}}, \ldots, x_{i_{m, K_m}}}$.
In the original $MNK$-landscapes \cite{aguirre2007}, these numbers are randomly and independently drawn from $[0, 1)$.
As a consequence, it is very unlikely that two different solutions map to the same point in the objective space.

\section{$\rho MNK$-Landscapes: Multiobjective $NK$-Landscapes with Correlation}
\label{sec:mnk}

In this section,
we define the $CMNK$- and the $\rho MNK$-landscapes, which are based on the $MNK$-landscapes \cite{aguirre2007}.
In this multiobjective model, the correlation between objective functions can be precisely tuned by a correlation matrix.
It allows to study the simultaneous influence of objective space dimension, non-linearity and objective correlation 
on the main properties of multiobjective fitness landscapes.
The construction of landscapes is defined and the analytic proof of the correlation between objectives, completed with an experimental study, are given.
Note that the proposed approach to tune the objective correlation can be applied to other MoCO problems where the objective functions
are summing objectives, share the same definition, but are computed with different cost or profit matrices.
This is the case, for instance, of the multiobjective knapsack, traveling salesman and quadratic assignment problems \cite{knowles2003,ehrgott2005}.

\subsection{Definition}

In the proposed $CMNK$-landscapes, the epistasis structure is identical for all the objective functions:
$\forall m \in [1,M]$,  $K_{m}=K$
and
$\forall m \in [1,M]$, $\forall j \in [1,K_m]$, $i_{m, j} = i_{j}$.
The fitness components are not defined independently.
The numbers $(y^{1,i}_{x_i, x_{i_1}, \ldots, x_{i_{K}}}, \ldots, y^{M,i}_{x_i, x_{i_1}, \ldots, x_{i_{K}}})$ follow a multivariate uniform law of dimension~$M$,
defined by a correlation matrix $C$.
Thus, the $y$'s follow a multidimensional law with uniform marginals 
and the correlations between $y^{m, i}_{\ldots}$s are defined by the matrix $C$.
So, the four parameters of the family of $CMNK$-landscapes are 
($i$)~the number of objective functions $M$,
($ii$)~the length of the bit string $N$,
($iii$)~the number of epistatic links $K$, 
and ($iv$)~the correlation matrix $C$.

The matrix $C$ is a symmetric positive-definite matrix where $\frac{M(M-1)}{2}$ numbers can be defined.
In order to limit the number of free numbers in matrix $C$,
we define the matrix $C_{\rho}=(c_{np})$ which has the same correlation between all the objectives:
$c_{nn} = 1$ for all $n$, and $c_{np} = \rho$ for all $n \not= p$.
In this case, we denote $CMNK$-landscapes by $\rho MNK$-landscapes,
and the original $MNK$-landscapes are equivalent to $\rho MNK$-landscapes with $\rho=0$.
However, 
it is not possible to have the matrix $C_{\rho}$ for all $\rho$ between $[-1,1]$.
$C_{\rho}$ must be positive-definite: $\forall u \in \Real^{M}$, $u^{t}C_{\rho}u \geq 0$.
So, $\rho$ must be greater than $\frac{-1}{M-1}$.
For two-objective problems, all the correlations between $[-1,1]$ are possible.
However, for three-objective problems, the correlation $\rho$ must fall in $[-0.5, 1]$.
Of course,
if one wants to study very negative correlations between some pairs of objectives,
it is possible to design a matrix~$C$ that keeps the condition that $C$ is positive-definite.

To generate random variables with uniform marginals and a specified correlation matrix $C$, we follow the work of Hotelling \cite{Hotelling36}.
We first generate $(Z_1,\ldots,Z_M)$ a multinormal laws of means $0$ and correlation matrix $R=2 \sin(\frac{\pi}{6}C)$.
Then, the values $z_i = \Phi(Z_i)$ are uniformly distributed with a correlation matrix $C$, 
where $\Phi$ is the univariate normal cumulative density function.
Note that this is not the only way to generate a multivariate uniform law.

\subsection{Correlation between Objective Functions}
\label{sec:corObjective}

The construction of $CMNK$-landscapes defines correlation between the $y$'s but not directly between the objectives. 
In this section, we prove by algebra that the correlation between objectives is tuned by the matrix $C$.
This proof is followed by an experimental analysis.

\paragraph{Theoretical analysis.}
Let $F_m = (f_{m NK}(x))$ be the fitness vector values of the $2^N$ solutions with respect to objective $m$. 
The correlation between objective $n$ and $p$ is:
$cor(F_{n}, F_{p}) = \frac{cov(F_{n}, F_{p})}{\sigma_n \sigma_p}$
where $\sigma_n$ and $\sigma_p$ are the standard deviations of fitness values over the landscape of the $n^{th}$ and $p^{th}$ $NK$ fitness functions.
$F_{n}$ (resp. $F_{p}$) corresponds to the average value of the $N$ vectors $F_{ni}$ (resp. $F_{pj}$) of fitness component values:
$$cov(F_{n}, F_{p}) =\frac{1}{N^2} \sum_{i,j=1}^{N} cov(F_{ni}, F_{pj})$$
By definition, 
when $i \not= j$, $cov(F_{ni}, F_{pj})=0$ and $cov(F_{ni}, F_{pi}) = c_{np} \cdot \sigma_{ni} \cdot \sigma_{pi}$,
where $c_{np}$ is the correlation defined in the matrix $C$, 
and $\sigma_{ni}$ (resp. $\sigma_{pi}$) is the standard deviation of fitness component $i$.
The correlation between objectives $n$ and $p$ becomes:
$$cor(F_{n}, F_{p}) = c_{np} \frac{\sum_{i=1}^{N} \sigma_{ni} \sigma_{pi}}{N^2\sigma_n \sigma_p}$$
By construction of the fitness functions, 
the following relation between standard deviations stands $\sigma_{n}^2 = \frac{1}{N} \sum_{i=1}^{N} \sigma_{ni}^2$ (resp. for $\sigma_{p}^2$).
On average, the $\sigma_{ni}$ are equal to the standard deviation of the uniform law on $[0,1)$.
\begin{equation}
\label{eq:cor}
E(cor(F_{n}, F_{p})) = c_{np}
\end{equation}
Then, the average of the correlations between objective functions are given by the matrix $C$.
In the $\rho MNK$-landscapes, the parameter $\rho$ allows to tune very precisely the correlation between all pairs of objectives.

\begin{figure} [t]
\begin{center}
\begin{tabular}{cc}
\includegraphics[width=0.5\textwidth]{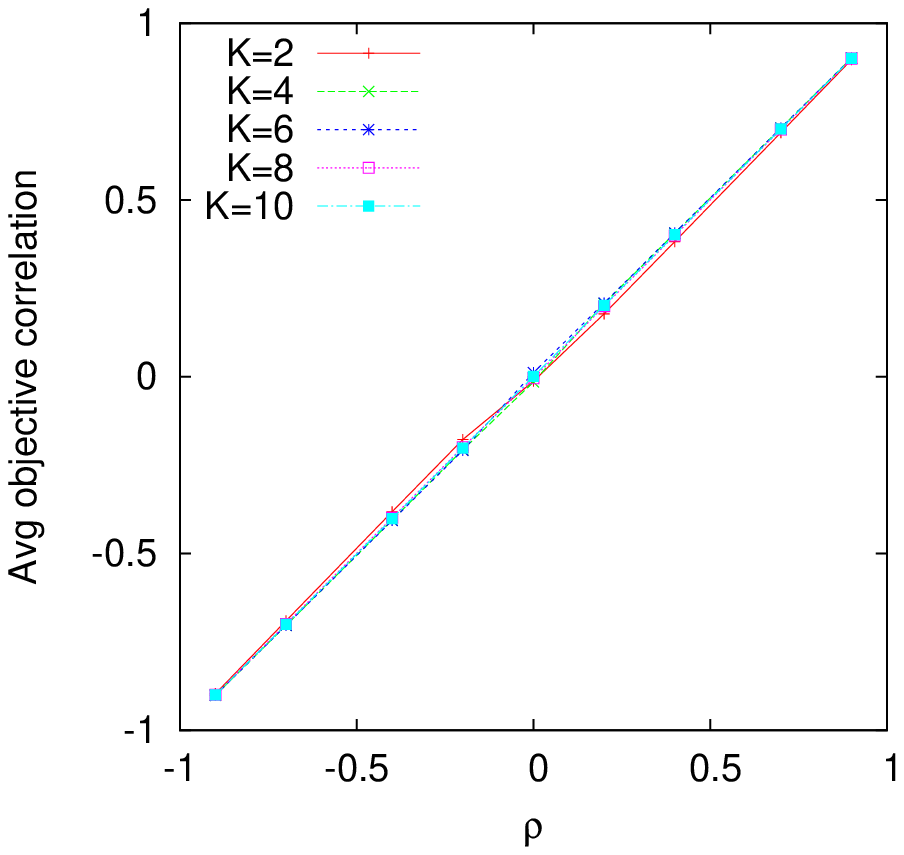} & 
\includegraphics[width=0.5\textwidth]{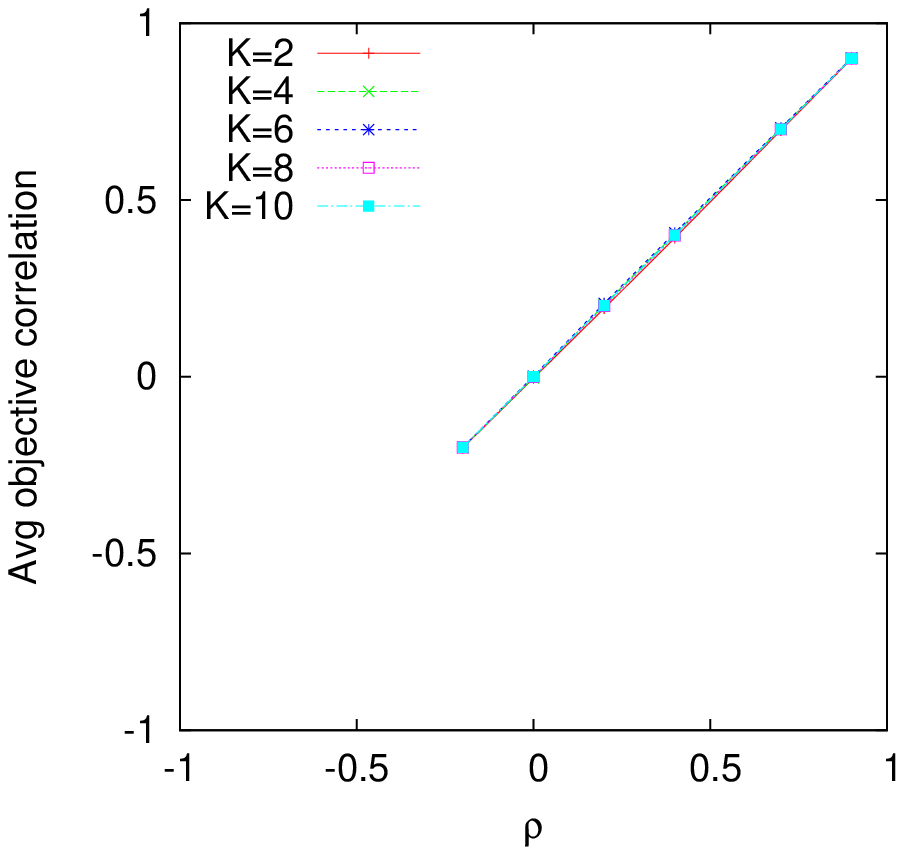} \\
\end{tabular}
\caption{Average values of the correlation between objectives according to the parameter~$\rho$.
The number of objectives is $M=2$ (left) and $M=5$ (right). \label{fig:objcor}}
\end{center}
\end{figure}

\paragraph{Experimental study.}
In order to enumerate the search space exhaustively, we conduct an empirical study for $N=18$.
In order to minimize the influence of the random creation of landscapes, 
we considered $30$ different and independent landscapes for each parameter combinations: $\rho$, $M$, $N$ and $K$.
The measures reported are the average over these 30 landscapes. 
The remaining set of parameters are given in Table~\ref{tab:param}.
%
Figure \ref{fig:objcor} shows the average%
\footnote{
For $M>2$, there are several correlation coefficients.
We report here the average correlation coefficients over all the objectives (these values are all very close).
}
of the Spearman correlation coefficient according to the parameters $\rho$, $M$ and $K$.
This confirms the result of equation~(\ref{eq:cor}),
the correlation coefficients are very close to the expected value $\rho$.

\begin{table}[t]
\caption{Parameters used in the paper for the experimental analysis.}
\vspace{-0.5cm}
\begin{center}
\begin{tabular}{c|l}
Parameter & Values \\
\hline
$N$ & 18 \\
$M$ & $\{2, 3, 5 \}$ \\
$K$ & $\{2, 4, 6, 8, 10 \}$\\
$\rho$ &  $\{ -0.9, -0.7, -0.4, -0.2, 0.0, 0.2, 0.4, 0.7, 0.9 \}$  such that $\rho \geq \frac{-1}{M-1}$ \\
\end{tabular}
\end{center}
\label{tab:param}
\end{table}%

Then, in the $\rho MNK$-landscapes, the parameter $\rho$ tunes very precisely the correlation, 
and, in addition to the correlated multiobjective quadratic assignment problem \cite{knowles2003}, it is possible to tune this correlation between all pairs of objectives.
In the following, we study the influence of epistasis, number of objective and objective correlation
on the properties of the efficient set for the $\rho MNK$-landscapes model.

\section{Analysis of the Efficient Set Properties}

In this section, 
we conduct experiments on the $\rho MNK$-landscapes 
in order to study different properties of the efficient set: its cardinality, the number of supported solutions and connectedness-related features.
The instances under study are defined by the parameter setting given in Table \ref{tab:param}.

\subsection{Cardinality of the Efficient Set}
\label{sec:cardinality}
\begin{figure} [ht]
\begin{center}
\begin{footnotesize}
\begin{tabular}{cc}
\includegraphics[width=0.4\textwidth]{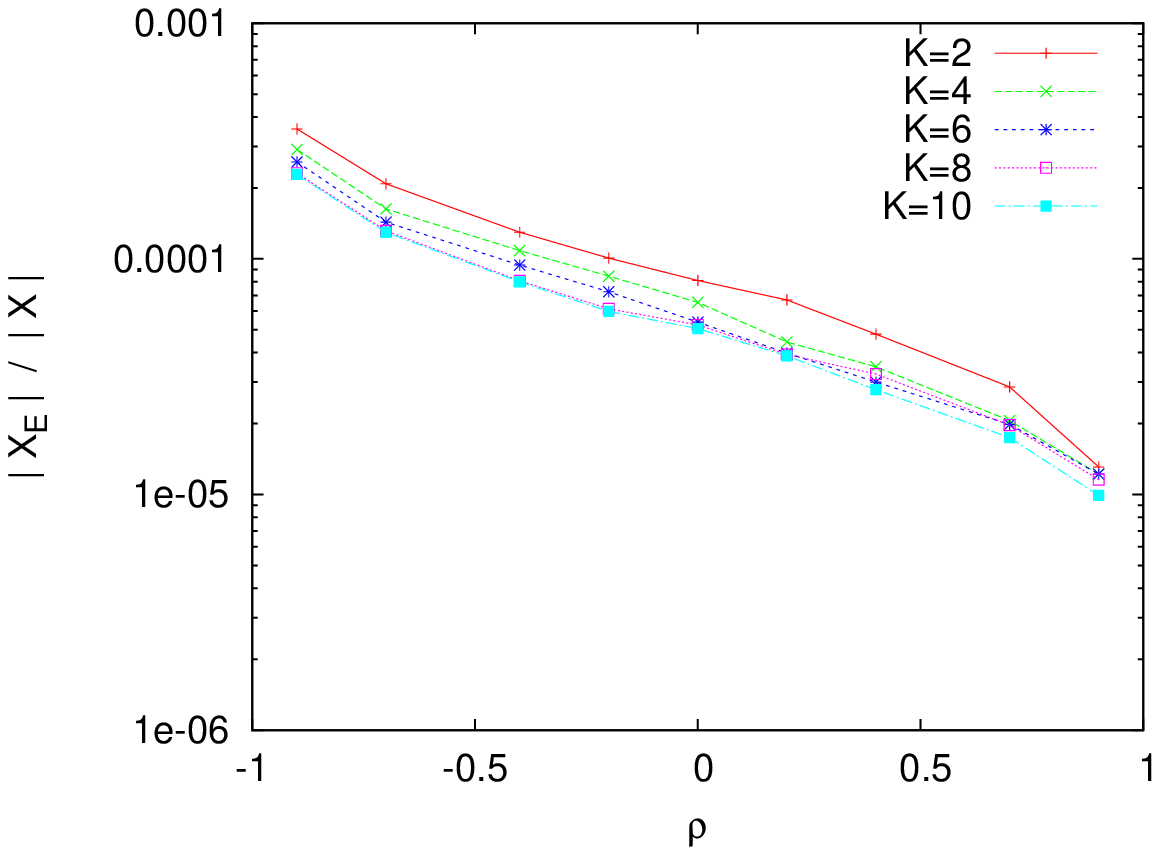} & \includegraphics[width=0.4\textwidth]{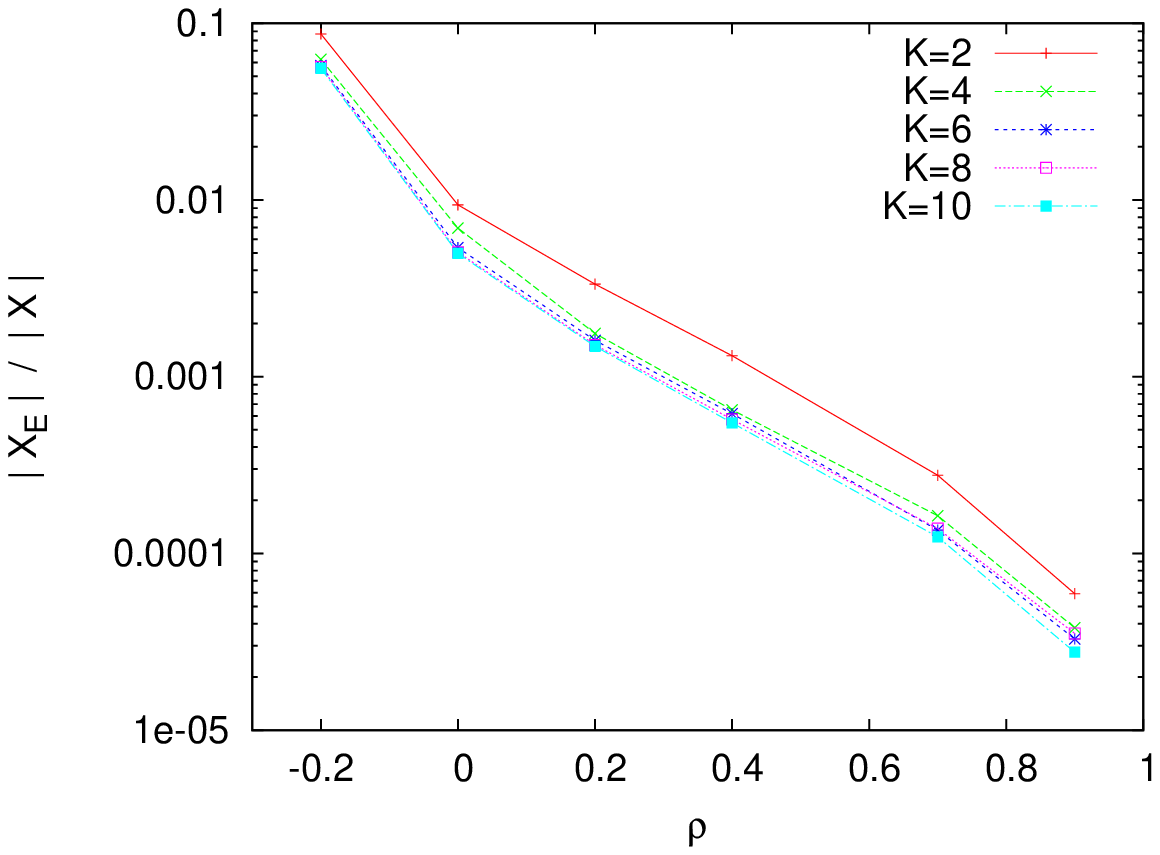} \\
\includegraphics[width=0.4\textwidth]{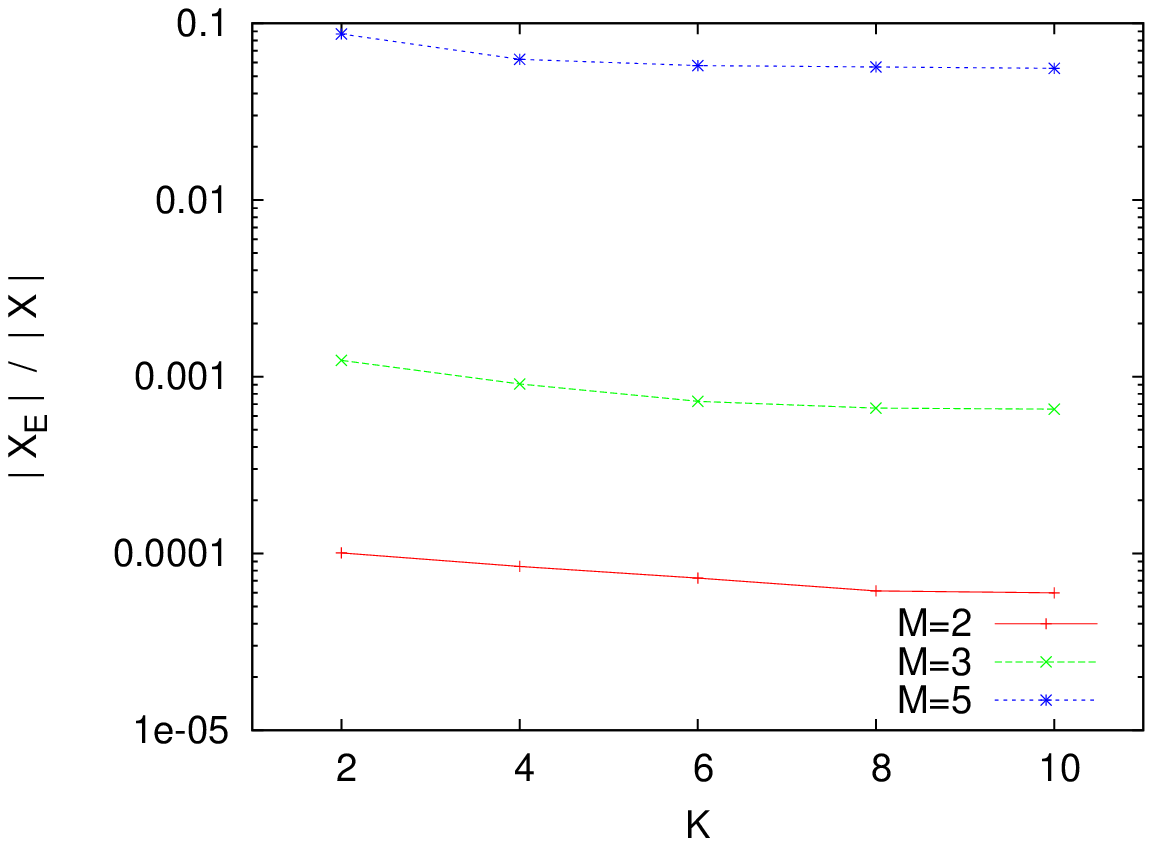} & \includegraphics[width=0.4\textwidth]{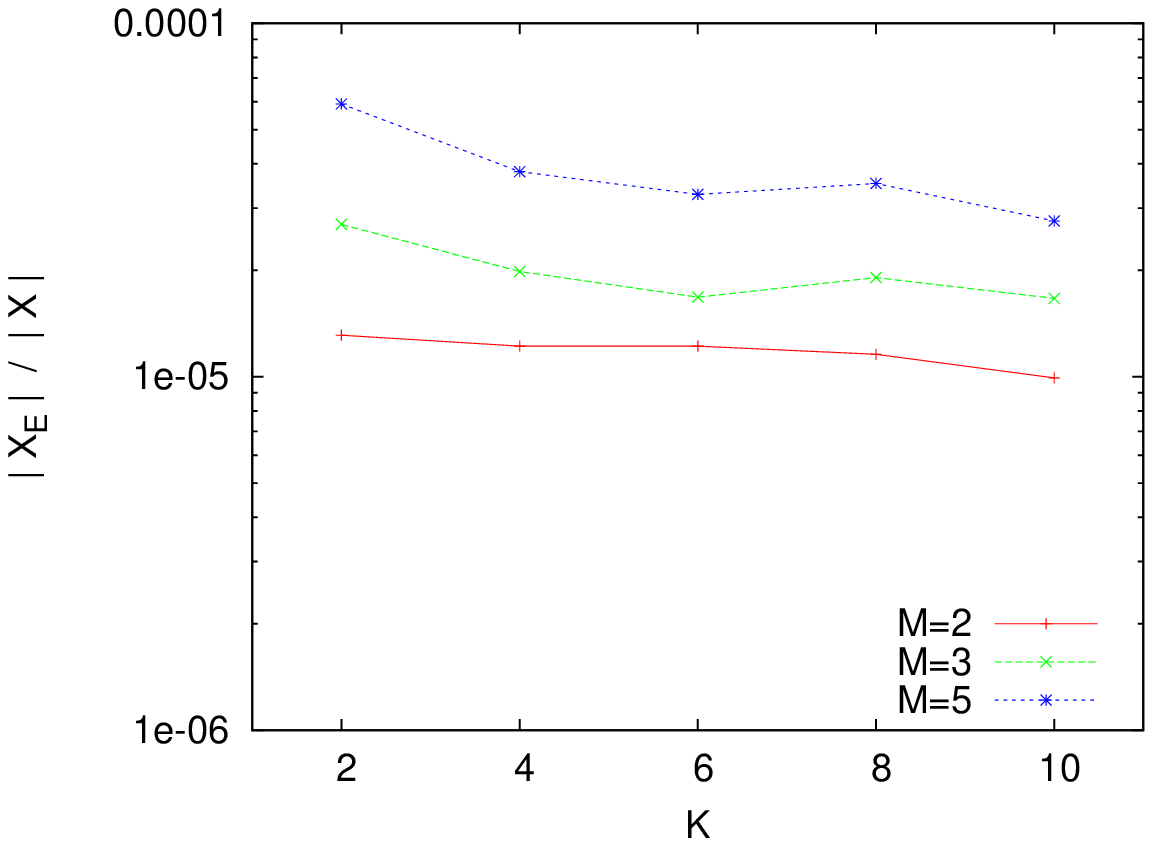} \\
\end{tabular}
\end{footnotesize}
\caption{Average ratio of the number of efficient solutions compared to the size of the search space ($2^N$)
according to parameter $\rho$ (top left $M=2$, right $M=5$),
and according to parameter $K$ for different number of objectives (bottom left $\rho=-0.2$, right $\rho=0.9$). 
Notice the log y-scale.
\label{fig:size_localPF}}
\end{center}
\end{figure}

Figure \ref{fig:size_localPF} shows the proportion of efficient solutions in the search space according to parameters $K$, $\rho$ and $M$ of $\rho MNK$-landscapes.
First of all, the epistatic parameter $K$ does not seem to have a major influence on the results.
At the opposite, the objective correlation $\rho$ modifies the number of efficient solutions to several orders of magnitude.
Indeed, the proportion decreases from $10^{-4}$ to $10^{-5}$ ($\rho \in [-1 , 1]$) for two-objective problems, 
and from $10^{-1}$ to $10^{-5}$ ($\rho \in [-0.2 , 1]$) for $M=5$.
With respect to the number of objective functions ($M=2,3$, and~$5$), the size increases of several decades according to $M$.
For a negative objective correlation ($\rho = -0.2$), the proportion goes from $10^{-4}$ to $10^{-1}$ 
whereas it goes from $10^{-5}$ to $10^{-4}$ for a positive correlation ($\rho = 0.9$).

The influence of objective correlation on the efficient size becomes as important as the number of dimension of objective space.
A lot of solutions becomes efficient when the anti-correlation is high.
Now, let us suppose that we want to set or to bound the size of the approximation set by $100$.
Such a parameter setting is often used while handling a population or an archive of non-dominated solutions in a multiobjective metaheuristic.
For the $\rho MNK$-landscapes,
the proportion of non-dominated solutions over the search space should be roughly around $4 \cdot 10^{-4}$
(this goes up to $8 \cdot 10^{-4}$ for $200$ solutions).
Whatever the correlation value $\rho$, a $100-$solution approximation set always allows to store all the efficient set for two-objective problems.
However, this is not the case for a higher dimension of the objective space.
For instance, for $M=5$, $100$ solutions suffice to store the whole efficient set for a high objective correlation only ($\rho > 0.5$).
In other words, for $\rho < 0.5$, we cannot pretend to identify the whole efficient set exhaustively by handling a $100-$solution approximation set.

To summarize, when the number of objective increases, and even more when the objectives are in conflict,
the size of the efficient set becomes very large, and then tend to be intractable.
In this case, it is not reasonable to pretend to identify the whole efficient set, and a limited-size approximation should be considered. 
This first result shows the importance to design a benchmark where the objective correlation can be tuned precisely, even when $M > 2$.
Such a property should be taken into consideration for the development of metaheuristics, when the number of objective becomes too large,
and when there is a high anti-correlation between objective functions.
A special attention should be paid with regards to the size of the approximation set handled by the search approach.

\subsection{Number of Supported Efficient Solutions}
\begin{figure} [ht]
\begin{center}
\begin{tabular}{cc}
\includegraphics[width=0.4\textwidth]{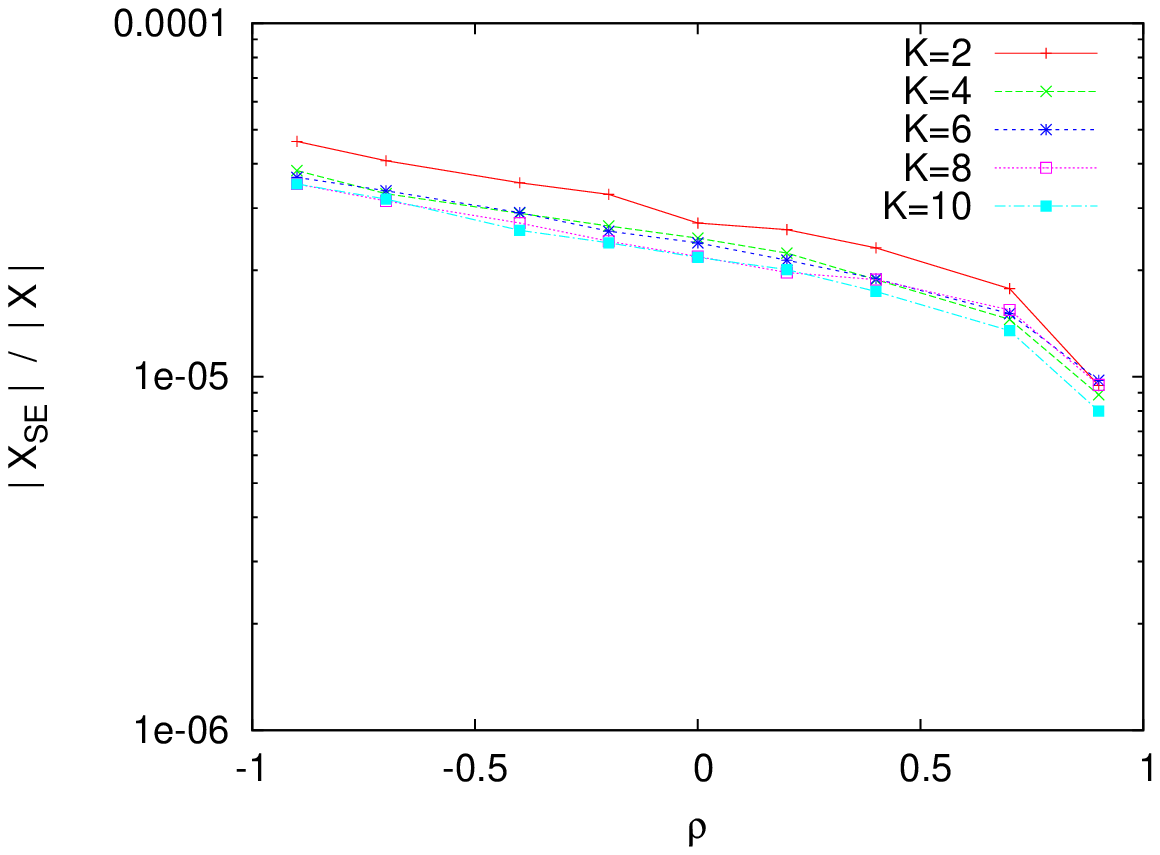} & \includegraphics[width=0.4\textwidth]{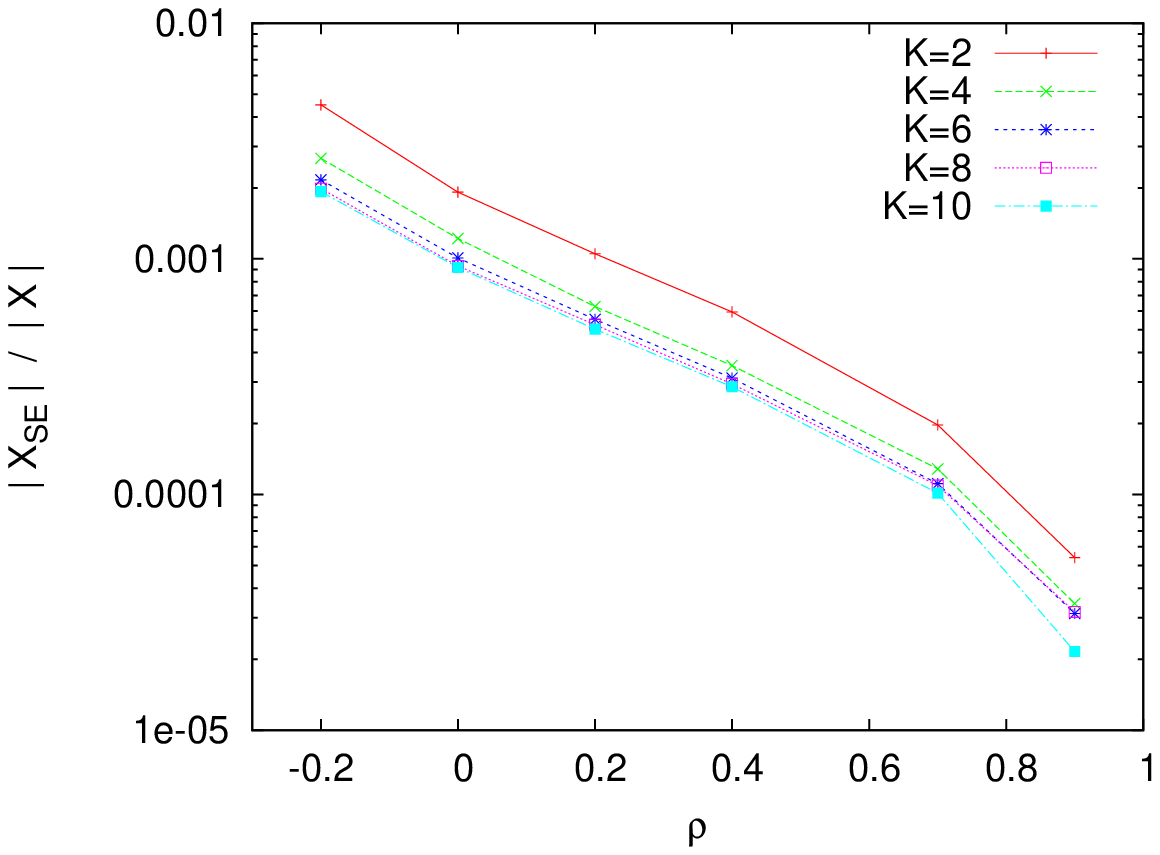} \\
\includegraphics[width=0.4\textwidth]{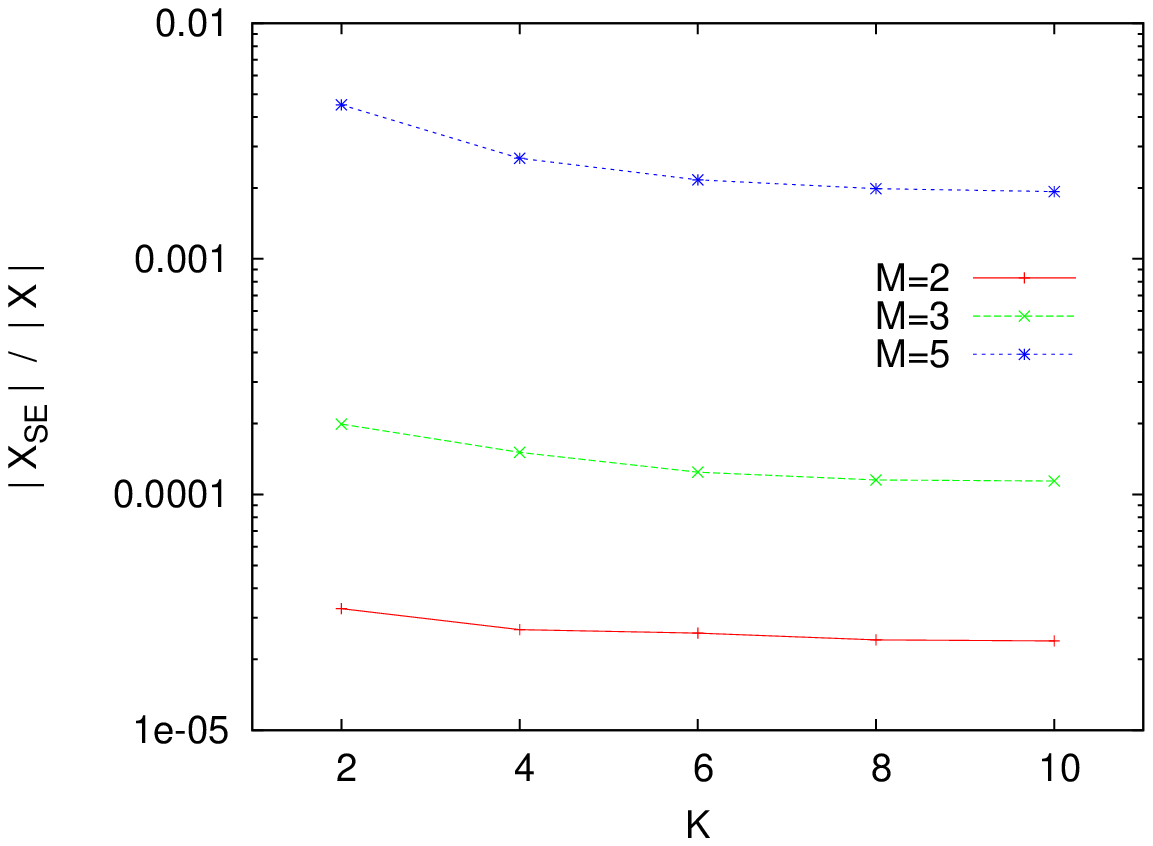} & \includegraphics[width=0.4\textwidth]{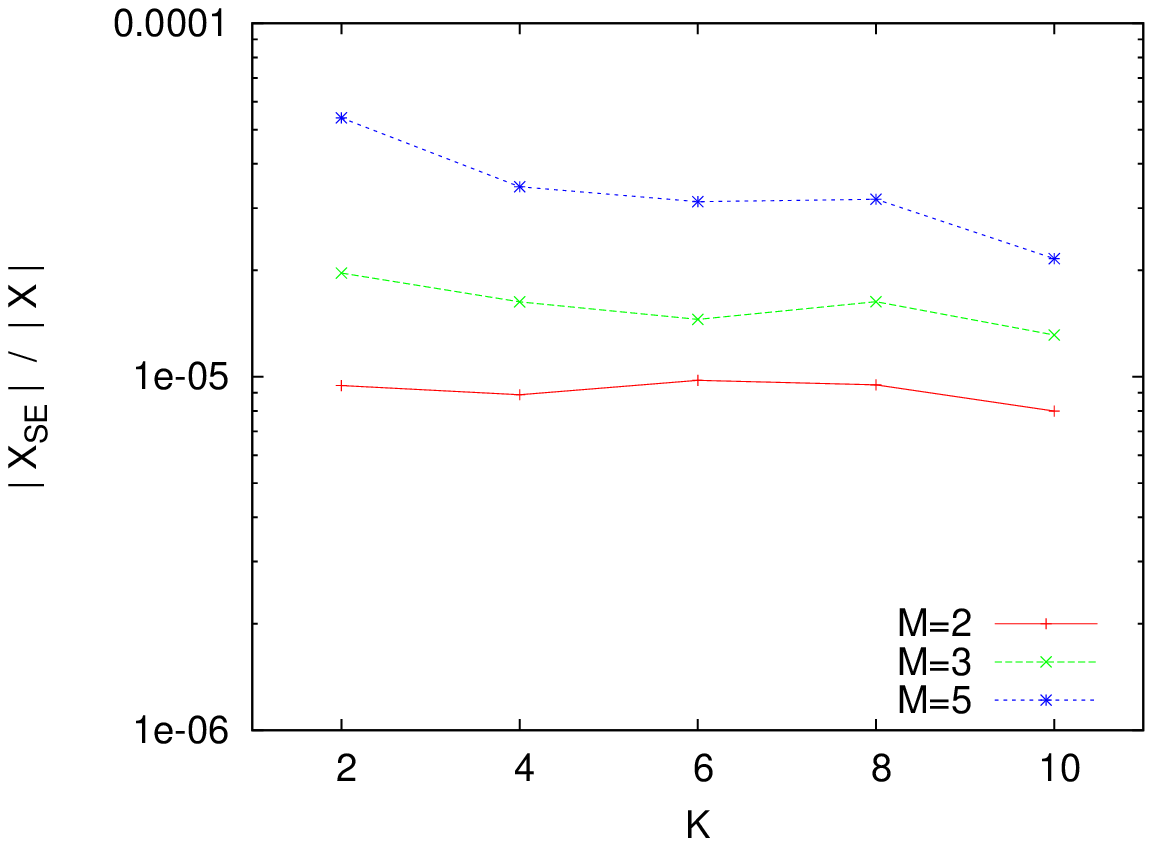} \\
\end{tabular}
\caption{Average ratio of the number of supported efficient solutions compared to the size of the search space ($2^N$)
according to parameter $\rho$ (top left $M=2$, right $M=5$),
and according to parameter $K$ for different number of objectives (bottom left $\rho=-0.2$, right $\rho=0.9$). 
Notice the log y-scale.
\label{fig:size_supported}}
\end{center}
\end{figure}

\begin{figure} [ht]
\begin{center}
\begin{tabular}{cc}
\includegraphics[width=0.4\textwidth]{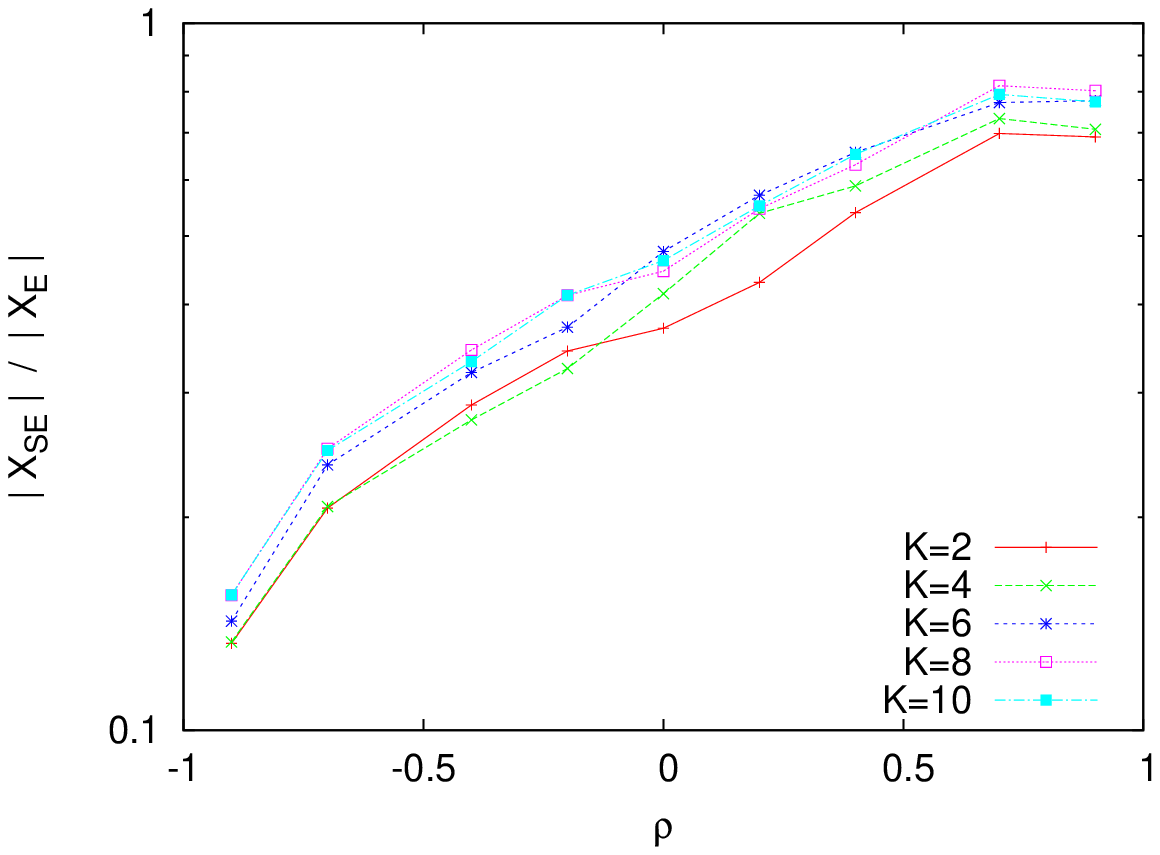} & \includegraphics[width=0.4\textwidth]{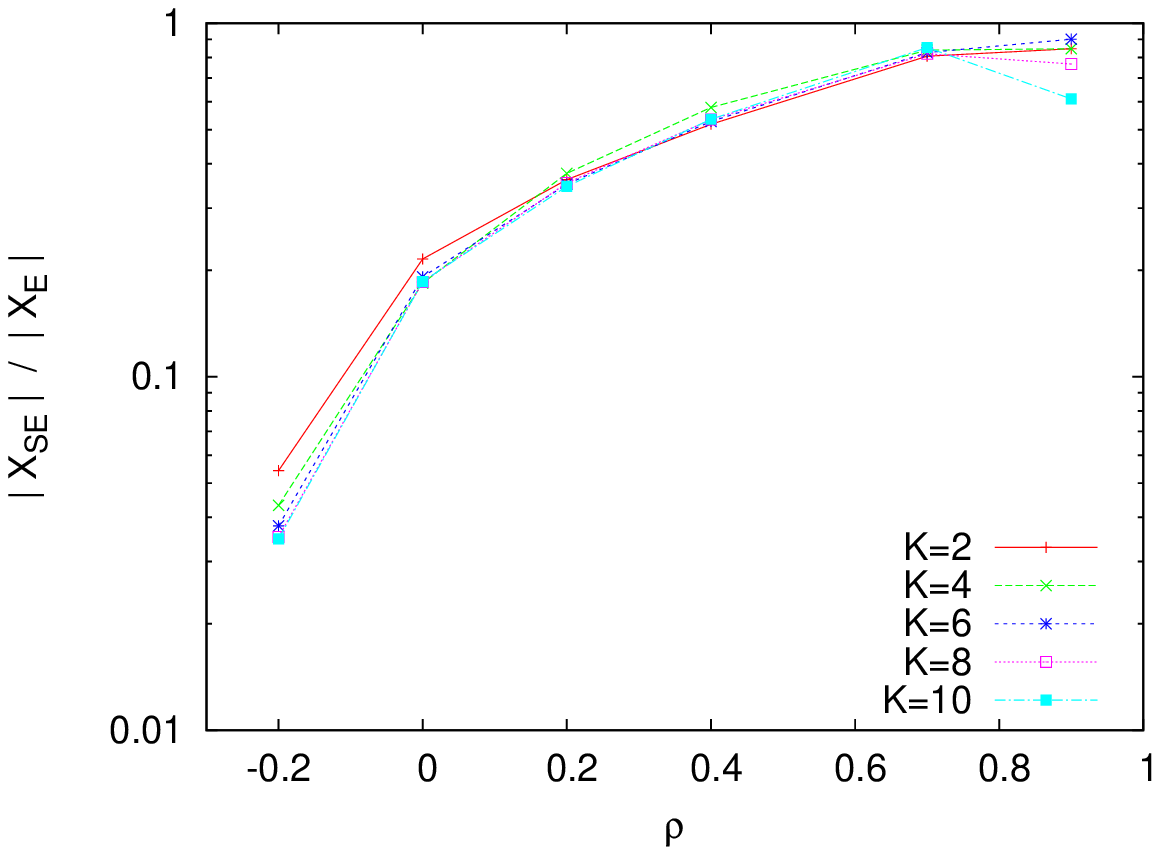} \\
\includegraphics[width=0.4\textwidth]{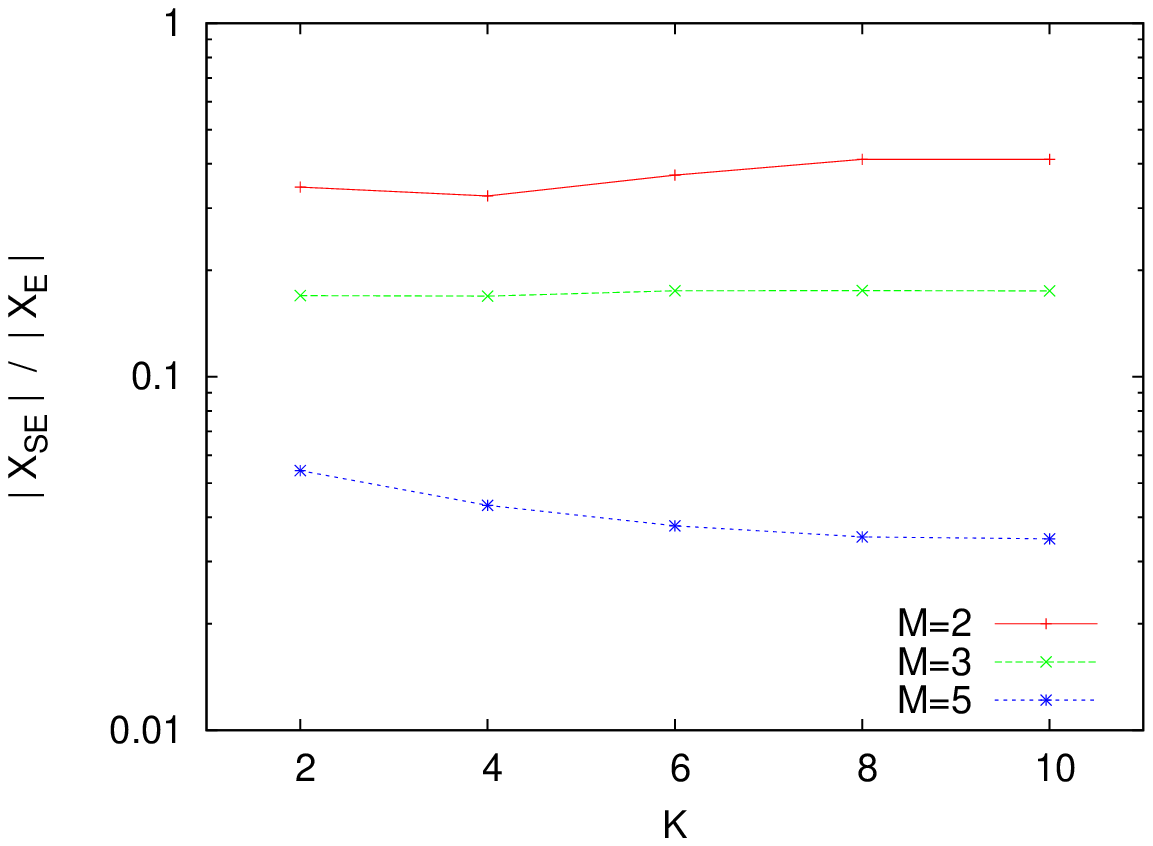} & \includegraphics[width=0.4\textwidth]{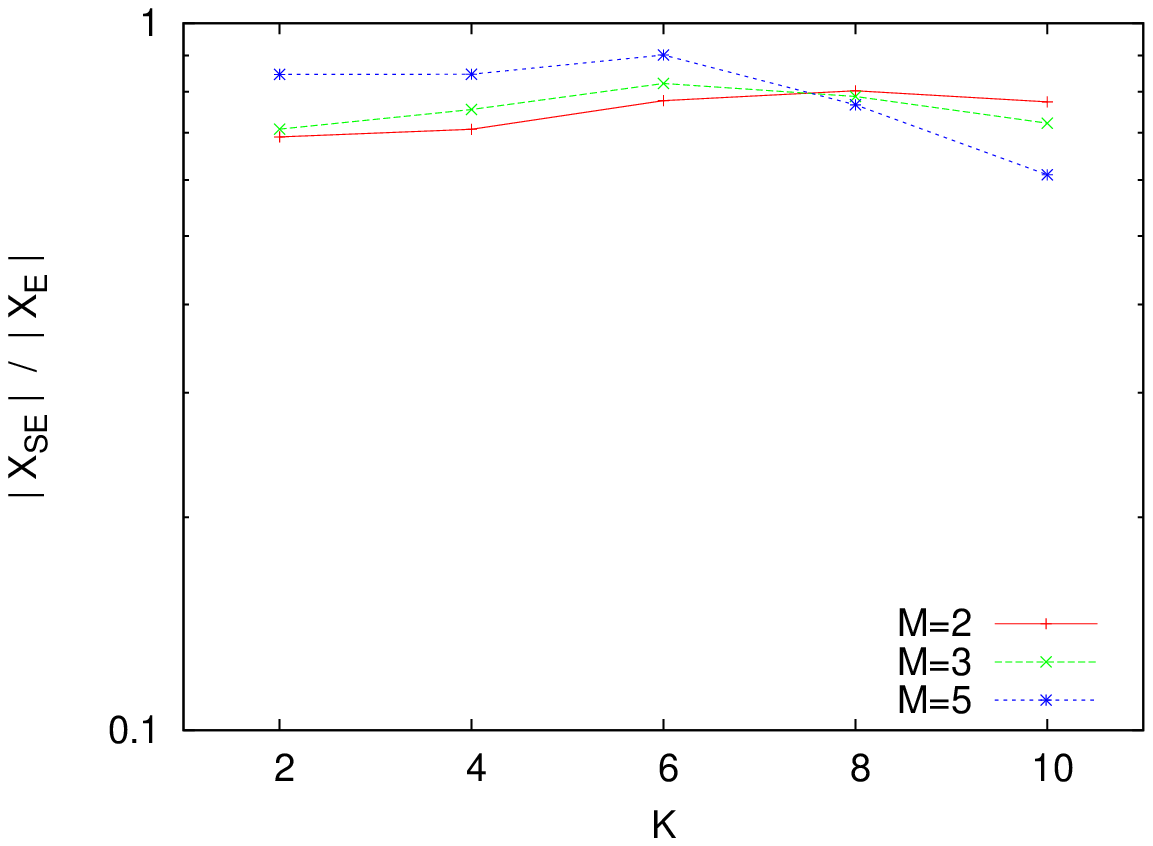} \\
\end{tabular}
\caption{Average ratio of the number of supported efficient solutions compared to the size of the efficient set
according to parameter $\rho$ (top left $M=2$, right $M=5$),
and according to parameter $K$ for different number of objectives (bottom left $\rho=-0.2$, right $\rho=0.9$). 
Notice the log y-scale.
\label{fig:size_relSupport}}
\end{center}
\end{figure}

Figure \ref{fig:size_supported} shows the proportion of supported solutions in the search space according to parameters $K$, $\rho$ and $M$ of $\rho MNK$-landscapes.
Mainly, this number follows the size of the efficient set: the epistatic parameters $K$ has low influence on the size.
When the objective space dimension increases or the objective correlation decreases, the number of supported solutions gets higher.
The difference with the size of the efficient set becomes more clear in Figure \ref{fig:size_relSupport}.
It gives the proportion of supported solutions over the efficient set. 
This proportion is nearly independent of the epistasis degree of the problem ($K$).
However, when the objective correlation increases, this proportion increases.
For a high objective correlation ($\rho=0.9$), nearly all solutions become supported (this is even the case for some instances).
The same observation can be made with the number of objectives. 
The number of supported solution increases with the cardinality of the efficient set, but the former increases faster than the latter.

While putting this property in relation with the design of a metaheuristic, 
we can conclude that scalar approaches should become more appropriate 
when the number of objective is low, and when the objective correlation is high.

\subsection{Connectedness of the Efficient Set}

In this section, the efficient graph (see Section \ref{sec:moco}), {\it i.e.} the graph of efficient solutions where edges are induced by a given neighborhood operator, is analyzed.

\begin{figure} [ht]
\begin{center}
\begin{tabular}{cc}
\includegraphics[width=0.4\textwidth]{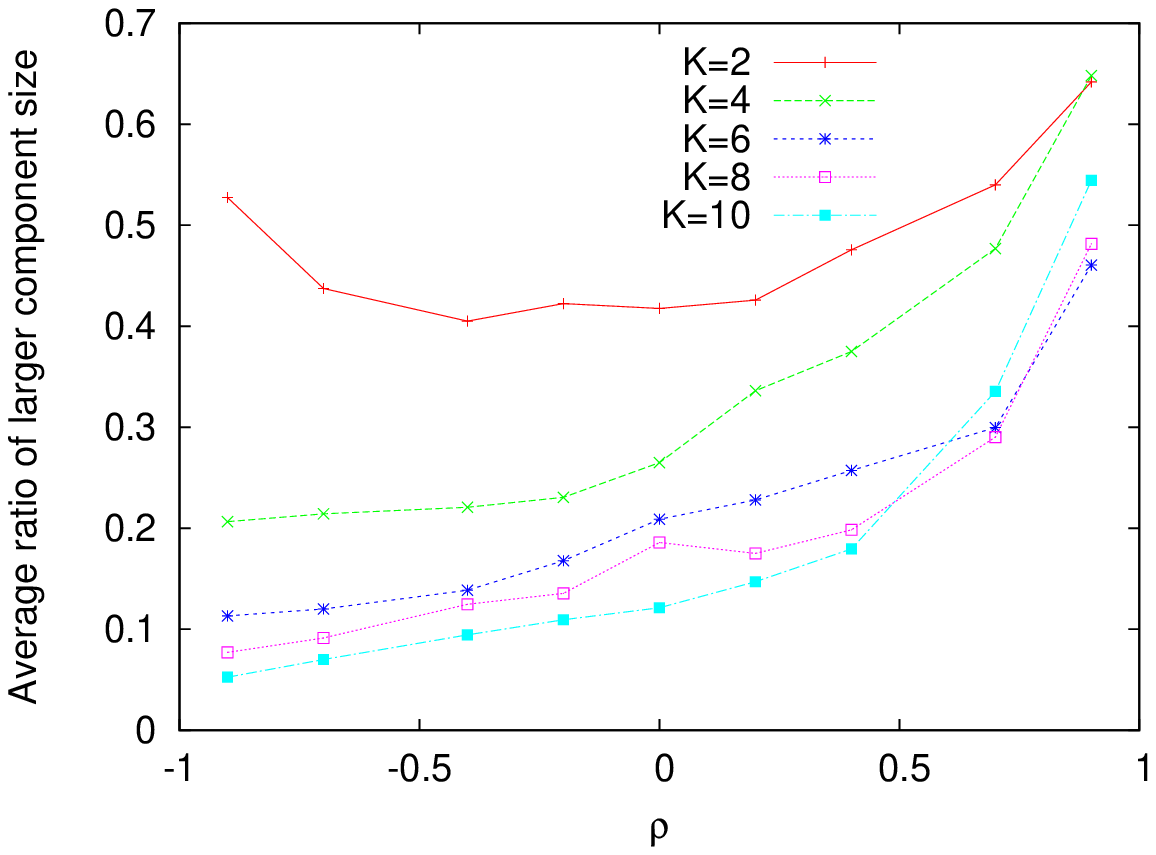} & \includegraphics[width=0.4\textwidth]{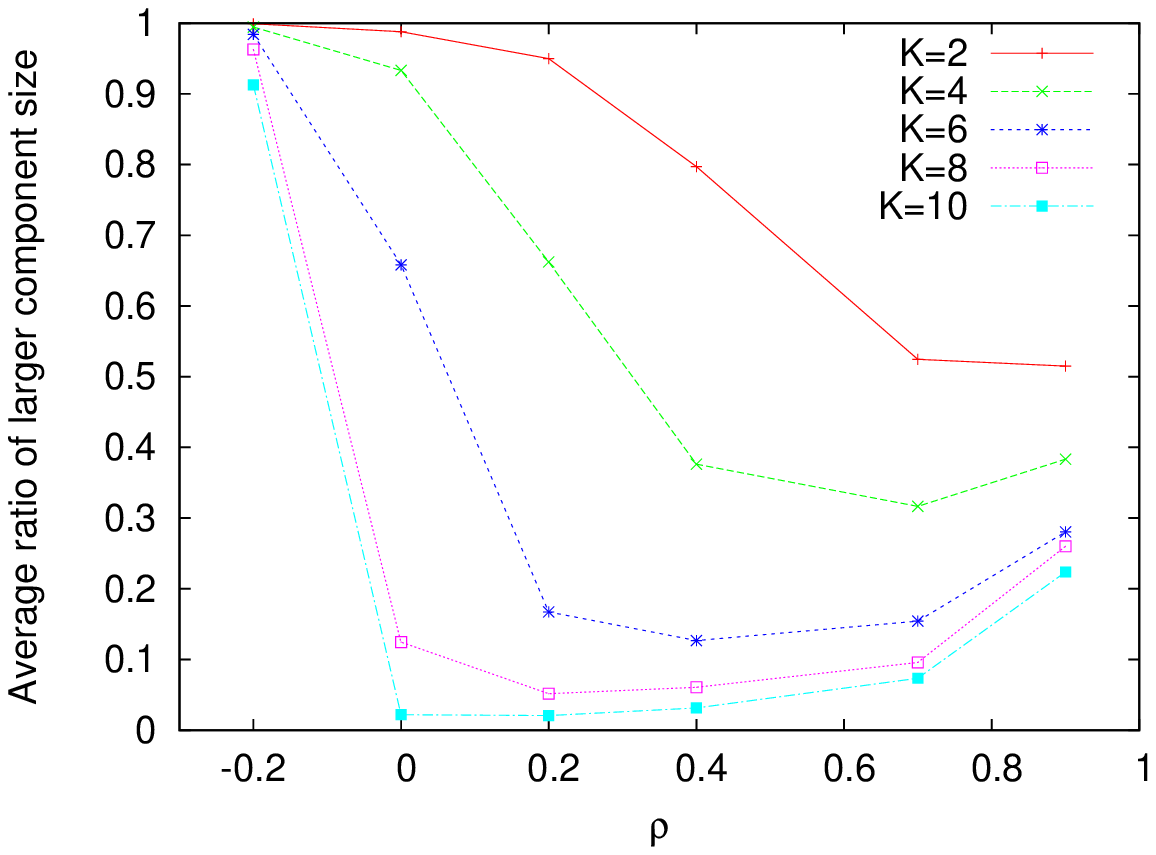} \\
\includegraphics[width=0.4\textwidth]{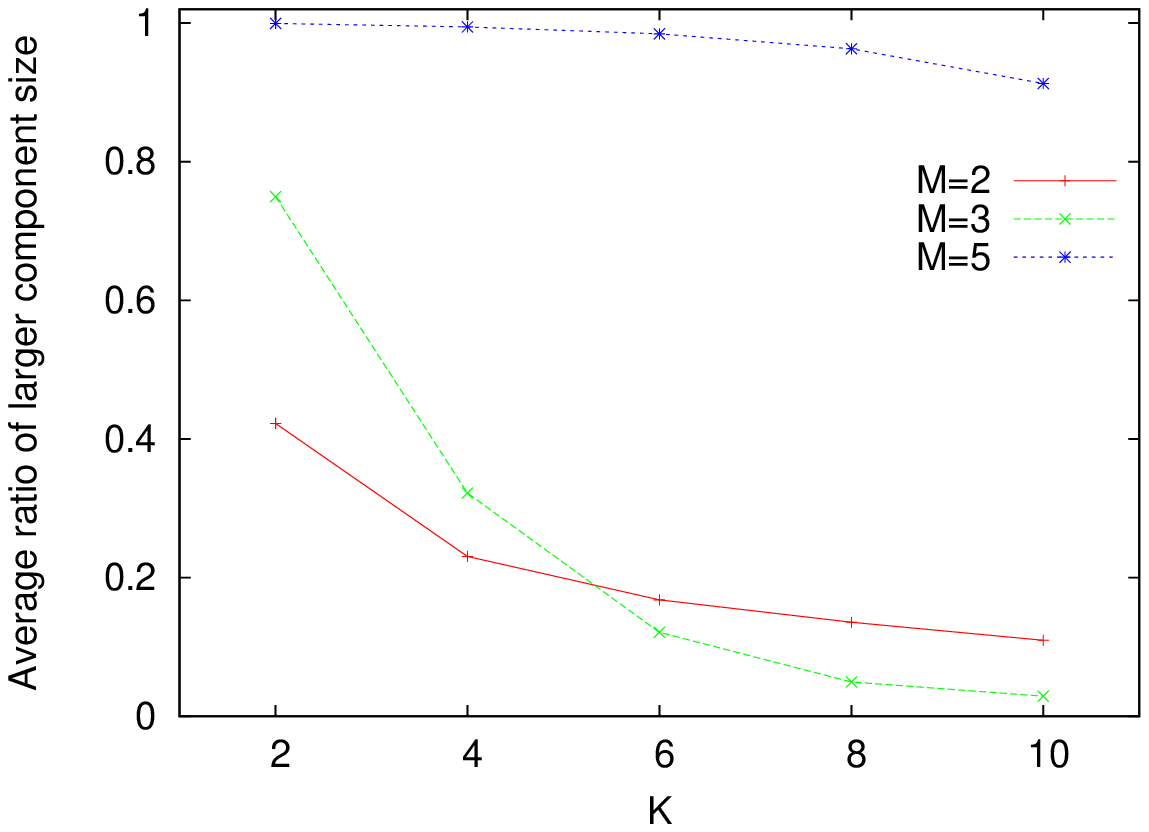} & \includegraphics[width=0.4\textwidth]{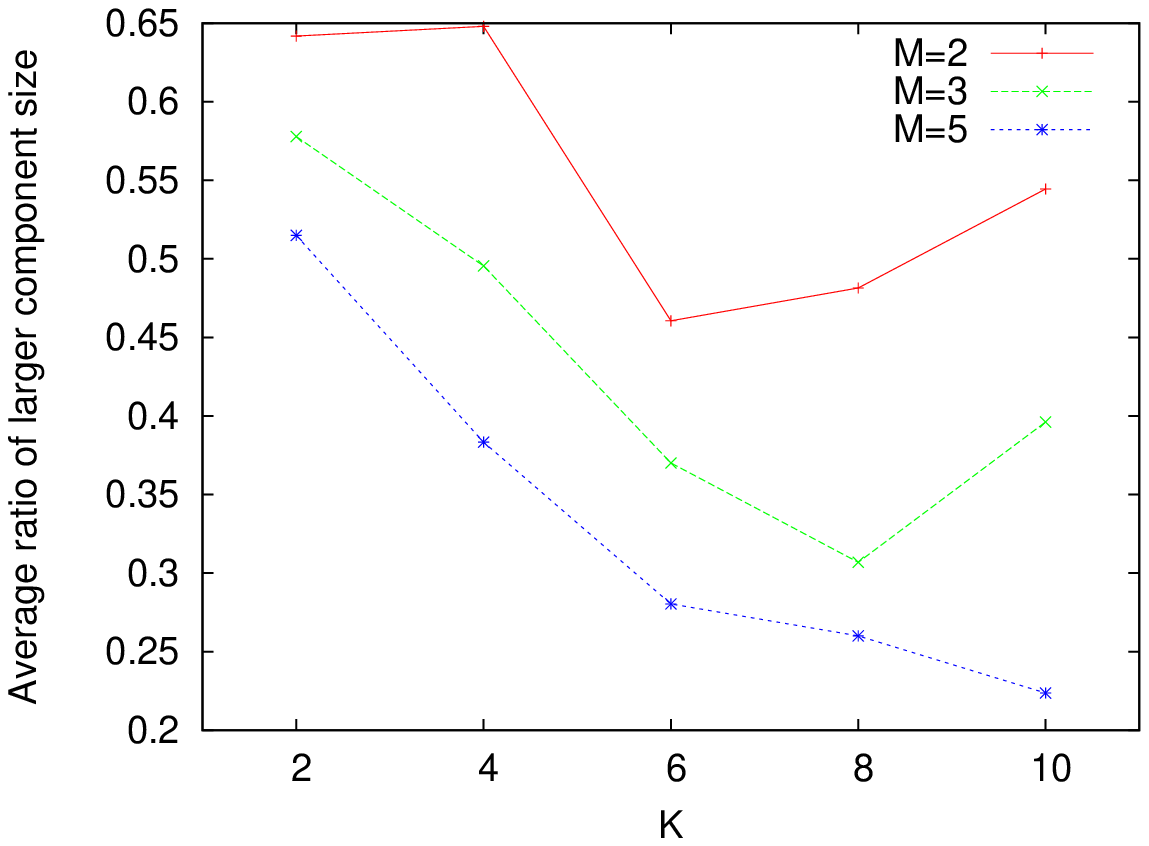} \\
\end{tabular}
\caption{Average ratio of the size of the larger component of the efficient graph and Hamming distance of 1 to the size of the efficient set 
according to parameter $\rho$ (top left $M=2$, right $M=5$),
and according to parameter $K$ for different number of objectives (bottom left $\rho=-0.2$, right $\rho=0.9$). 
\label{fig:graphPF_sizeLarger}}
\end{center}
\end{figure}

Firstly, the efficient graph can be composed of several connected components. 
In this case, all the efficient solutions are not connected with respect to the neighborhood relation.
Figure \ref{fig:graphPF_sizeLarger} shows the average ratio of the larger connected component size induced by Hamming distance $1$.
Nearby all solutions of the efficient graph are in the same component when the objective space dimension is high ($M=5$)
and when the objective correlation is negative ($\rho=-0.2$).
At first sight, such a result seems to be explained by the very large size of the efficient set obtained for those parameters (see Section \ref{sec:cardinality}).
However, we compared this result to the size of the larger component of a graph of same size, but where the nodes are now random solutions.
We found out that this size is much smaller than the one of the efficient graph, in particular when the epistatic degree is low 
($170$ times larger for $M=5$, $\rho=-0.2$, and $K=4$).
Consequently, the ratio size of the larger component is not the consequence of the number of efficient solutions only .

Contrary to the size of the efficient set, the size of the largest connected component seems to depend on the epistatic degree $K$.
Indeed, this size decreases when $K$ increases.
As an example, for $M=2$ and $\rho=-0.4$, the ratio size is $0.42$ for $K=2$ and lower than $0.1$ for $K=10$.
When the epistatic degree is low, the objective values of neighboring solutions are correlated, and this correlation decreases with the epistatic degree \cite{WEI:90}.
This could explain our experimental result:
If a solution is efficient, the probability that one of its neighbors is also efficient gets higher when the epistatic degree gets lower.

The objective correlation and the number of objective functions also affect the size of the largest connected component.
But the variation is different with respect to the number of objective functions.
For $M=2$, the ratio of the larger component size increases when the objective correlation increases (apart from $K=2$).
For $M=5$, the ratio decreases when the objective correlation increases.
As a consequence, excepting when the efficient set is intractable (that is, when there is a high objective space dimension and a high anti-correlation degree),
we cannot expect to reach all the efficient solutions by iteratively exploring the neighborhood of an approximation set initialized with one non-dominated solution. 
However, when there are several connected components for the efficient graph based on Hamming distance $1$
(see the definition of cluster in Section \ref{sec:moco}), the distance between those components could be small.

\begin{figure} [ht]
\begin{center}
\begin{tabular}{cc}
\includegraphics[width=0.4\textwidth]{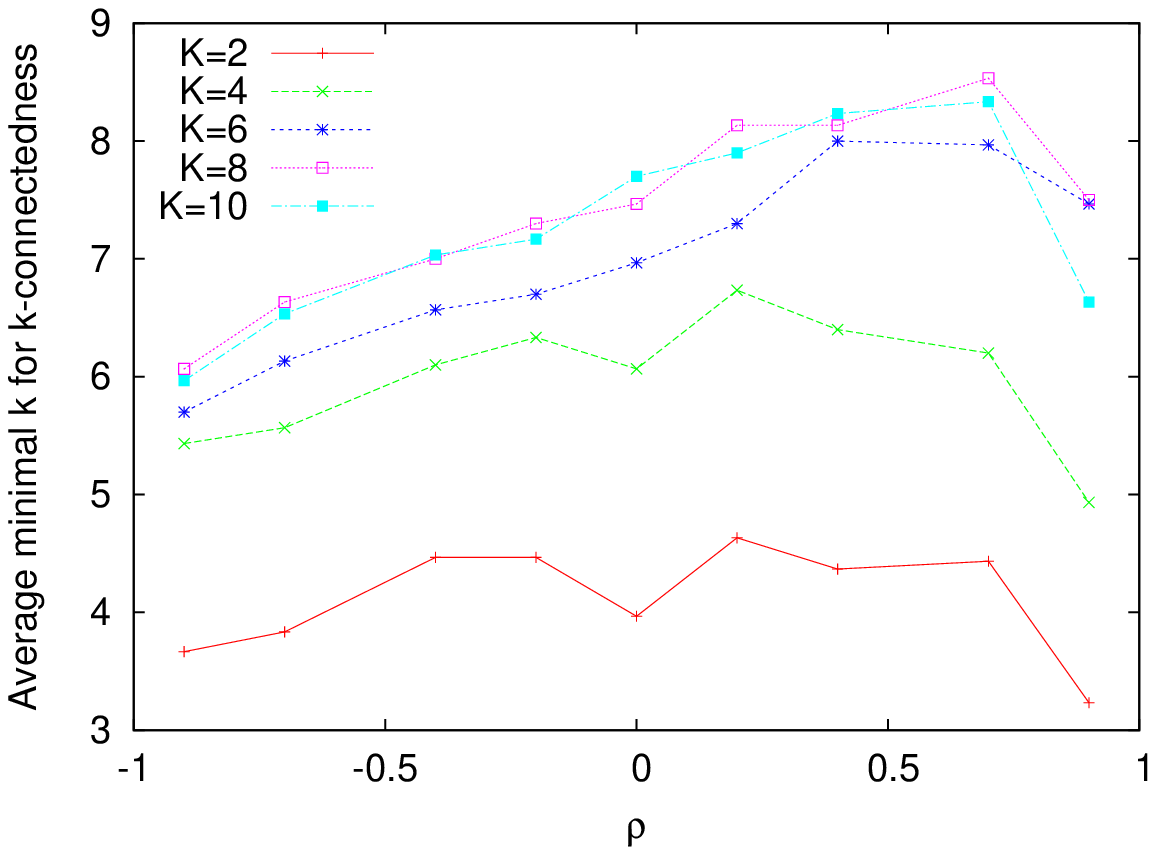} & \includegraphics[width=0.4\textwidth]{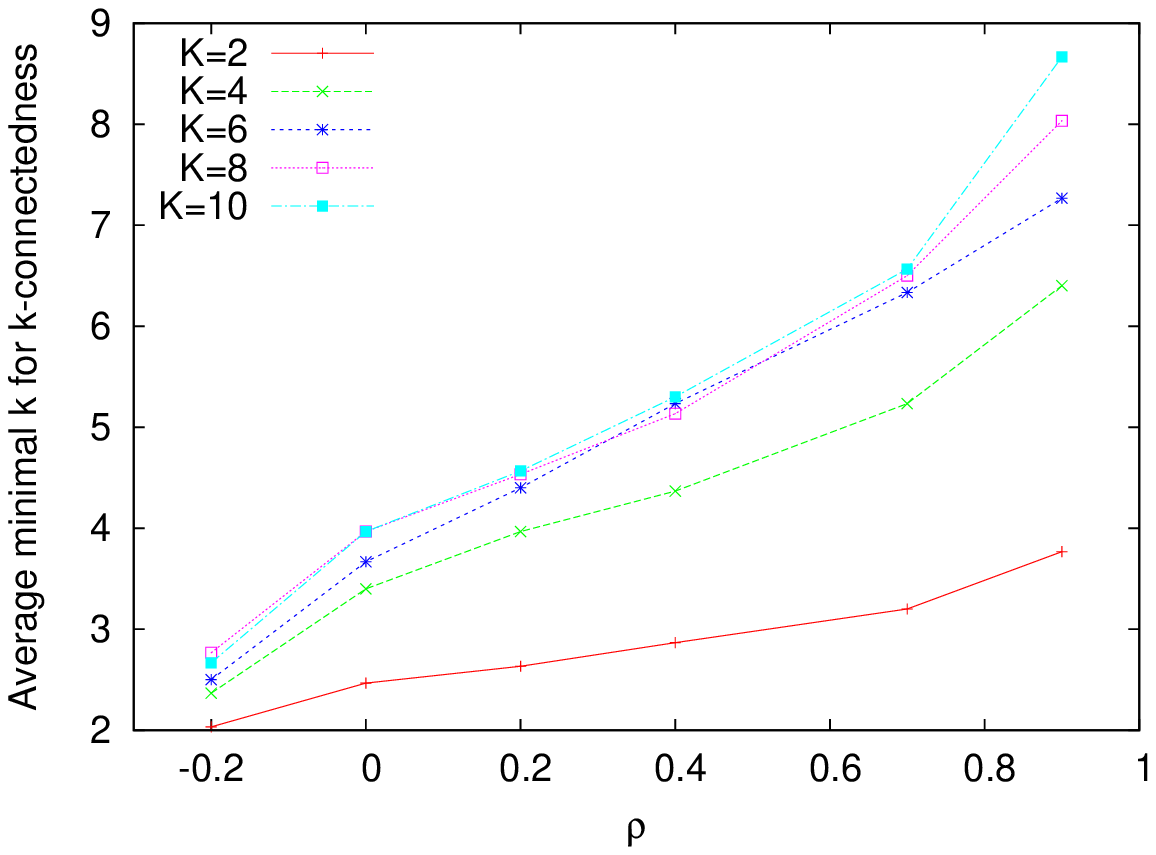} \\
\includegraphics[width=0.4\textwidth]{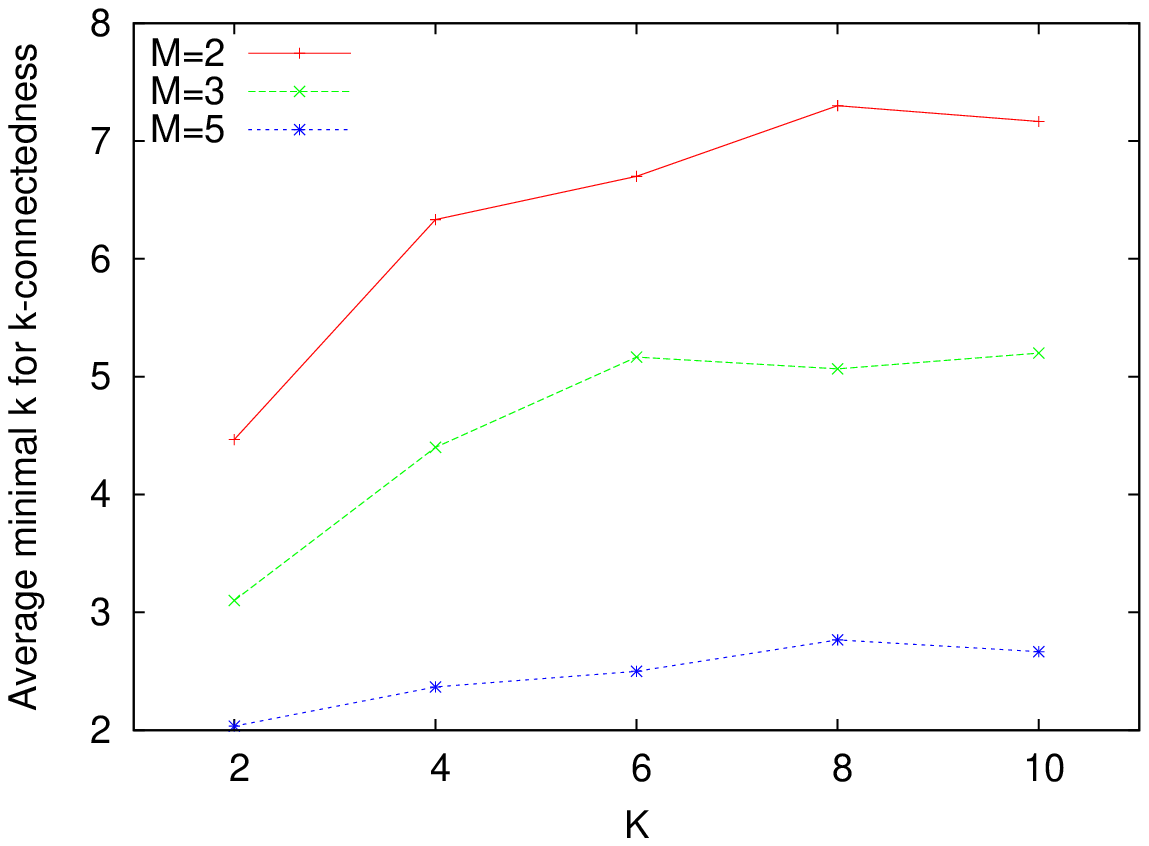} & \includegraphics[width=0.4\textwidth]{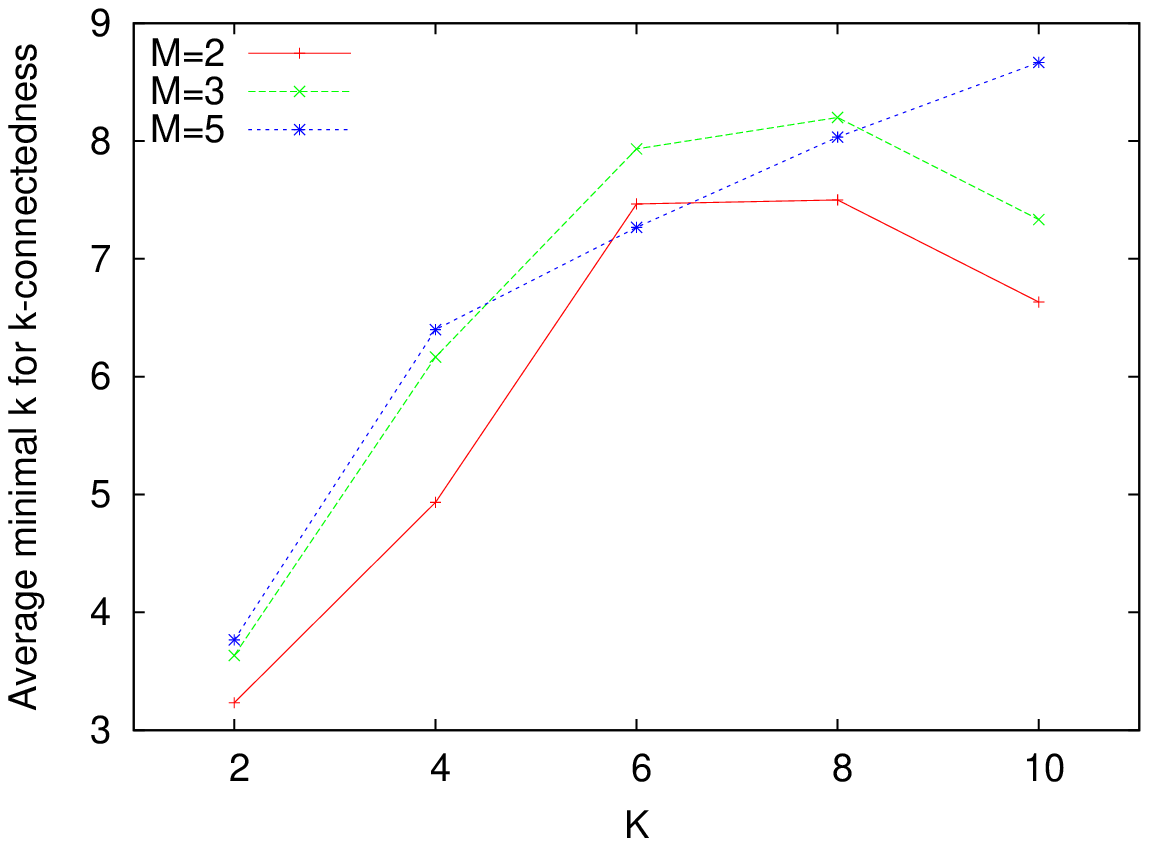}
\end{tabular}
\caption{Average of the minimal Hamming distance to connect all the efficient solutions
according to parameter $\rho$ (top left $M=2$, right $M=5$),
and according to parameter $K$ for different number of objectives (bottom left $\rho=-0.2$, right $\rho=0.9$). 
\label{fig:graphPF_Kconnect}}
\end{center}
\end{figure}

When efficient solutions are connected with respect to a neighborhood structure related to Hamming distance $k$ and not $k-1$,
the efficient set  is then said to be $k$-connected \cite{paquete2009}.
When the minimal distance $k$ is around $9$, which is the average distance between random solutions,
we can say that the distance between efficient solutions is large.
Figure \ref{fig:graphPF_Kconnect} shows the average minimal distance~$k$ to connect all the efficient solutions.
This minimal distance $k$ increases when the epistatic degree increases. 
As an example, for $\rho=-0.2$,
the average distance is equals to $4.3$ and $2$ for dimension 2 and 5, respectively, when $K=2$,
whereas it is equal to $7.1$ and $2.8$, respectively, when $K=10$.
These results meet the previous ones on the largest component size:
At the same time, the size of the larger component decreases, and the distance between efficient solutions increases.

The average $k$-connectedness increases also when the objective correlation increases.
For an objective space dimension $5$ and a negative objective correlation $\rho=-0.2$, 
it could be possible to reach all non-dominated solutions from another one, as the average minimal distance is lower than $3$.
At the opposite, when the objective correlation is positive,
it should be easier to find a new non-dominated solution by restarting the search from a random solution,
rather than exploring the neighborhood of a given non-dominated solution such as the distance is around the third of the bit string length.
When objectives are correlated, less solutions are to be found, but knowing some of them will not help to find more.
Then, the design of an efficient metaheuristic has to be different according to the objective correlation.
In a \emph{two-phase} approach,
the number of starting solutions and the size of the neighborhood can be tuned according to correlation between objectives
following this study.

\section{Discussion}
\label{sec:discs}
In this paper, we analyzed the consequence of the objective space dimension, the non-linearity, and the objective correlation
on the structure of multiobjective combinatorial search spaces for the design of metaheuristics.
We proposed a new method to design a multiobjective combinatorial benchmark where the correlation between all pairs of objectives can be tuned very precisely. As an example, we defined the $\rho MNK$-landscapes which extend the multiobjective $NK$-landscapes.

\begin{figure} [!ht]
\begin{center}
\begin{tabular}{ccc}
\includegraphics[width=0.35\textwidth]{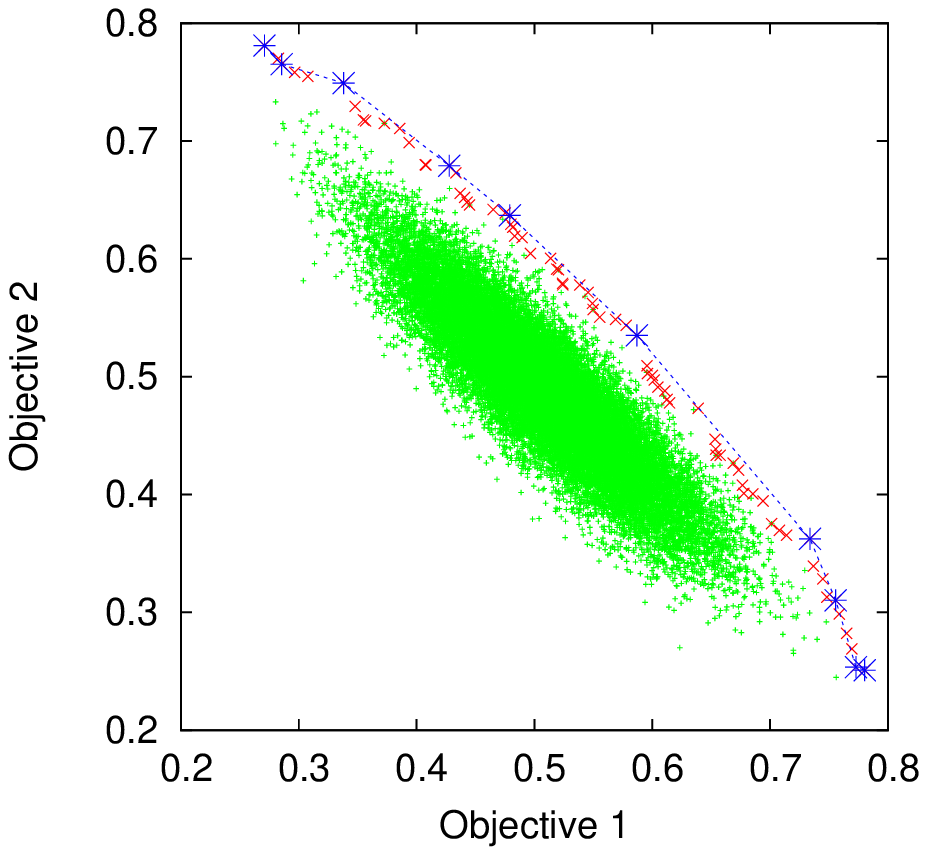} &
\includegraphics[width=0.35\textwidth]{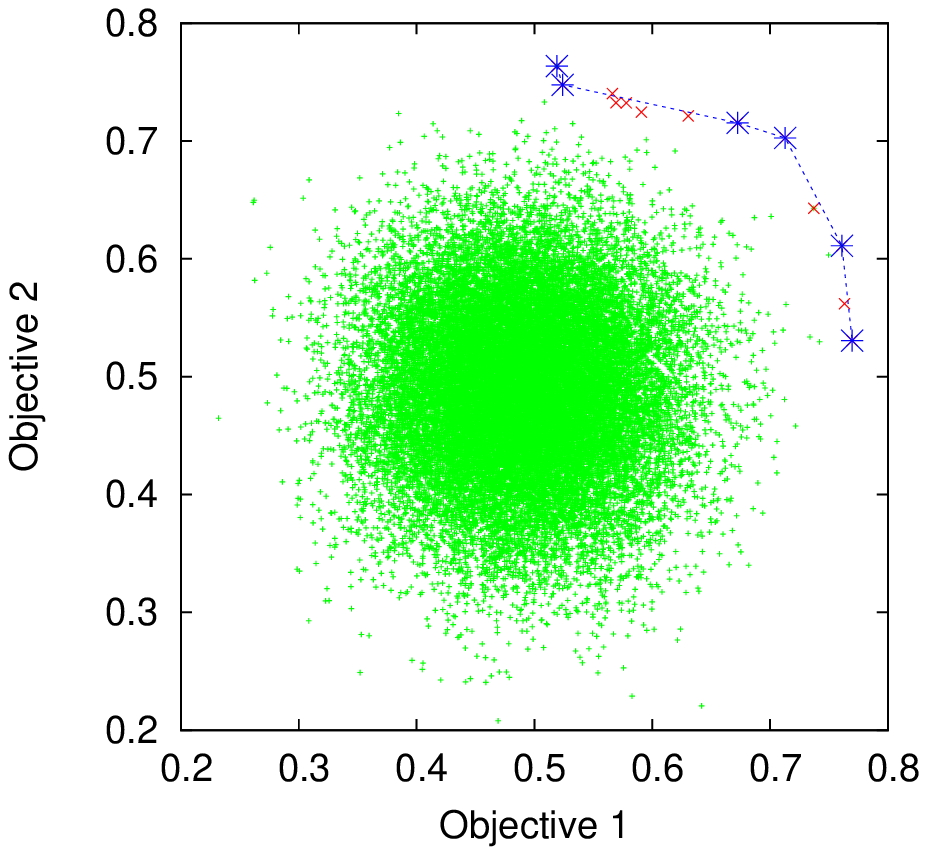} &
\includegraphics[width=0.35\textwidth]{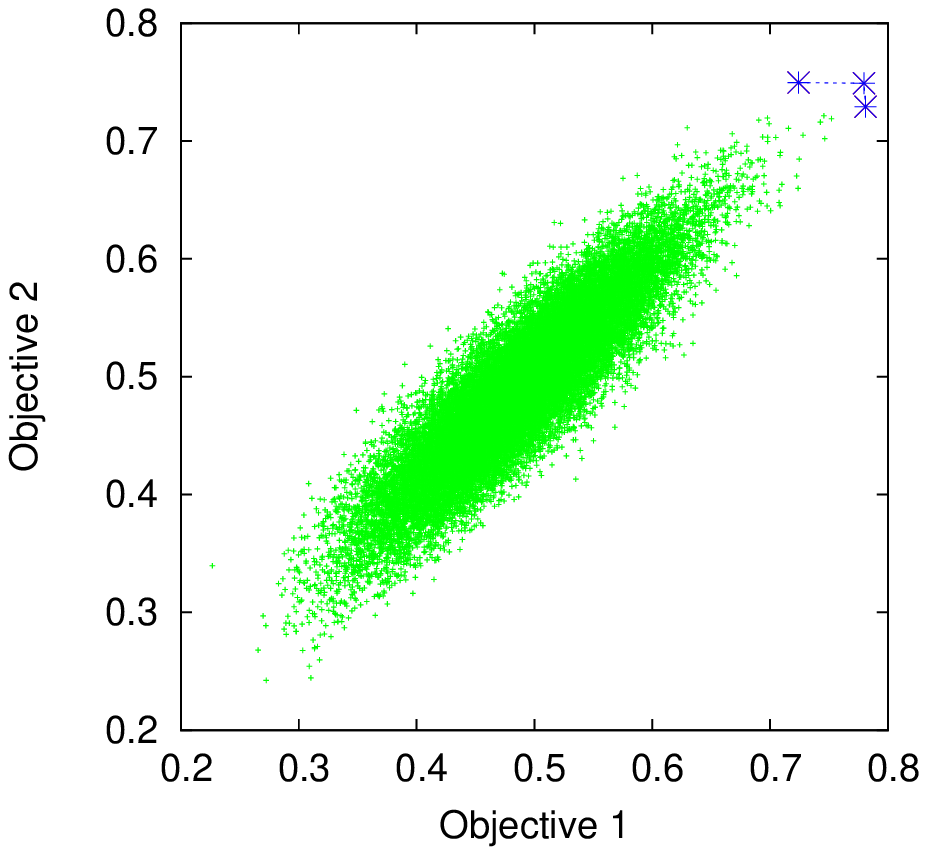} \\
\end{tabular}
\caption{
The objective space (maximization problem) for three landscapes. 
The number of objective is $M=2$, the length of bit string remains $N=18$, the epistasis parameters is $K=4$.
From left to right, the correlation increases from negative correlation to positive correlation ($\rho=-0.9, 0.0$ and $0.9$). 
The green points are random solutions of the search space ($10\%$ of the size), 
the red points are the solutions of the efficient set,
and blue are the supported solutions of the efficient set.
\label{fig:objSpaceExample}}
\end{center}
\end{figure}

Figure \ref{fig:objSpaceExample} shows three examples of $\rho MNK$-landscapes in the objective space.
The number of objective is $2$, the parameter $K$ is $4$, and length of the bit string is $18$. 
This gives a summary of our results in a more intuitive way.
When the objective correlation is negative, the objectives are in conflict (feasible solutions are in green). 
The efficient set size (in red) is large, and the problem could become intractable. 
In this case, a metaheuristic has to find a limited-size approximation of the efficient set only. 
When the objective correlation is null, as in \cite{aguirre2007},
the image of the search space in the objective space can be represented as a multidimensional `bowl'.
The objectives are independent.
When the objective correlation is positive,
there exists few solutions in the efficient set.
Nearly all solutions become supported.
Indeed, when the number of objectives is low, and when the objective correlation is high, efficient solutions are supported. 
We can conclude that scalar approaches should become more appropriate in such a case. 
The connectedness property is not represented in the last figure.
The size of larger connected component and the minimal distance to connect all the efficient solutions depend on the objective space dimension, the epistatic degree, and also on the objective correlation. 
A two-phase strategy, starting from some efficient (supported) solutions, and exploring their neighborhood at a given distance, 
can be tuned according to the results of this work.

Bringing those properties with the design of local search metaheuristics help to make proper choices between several classes of methodologies.
This analysis shows the importance of the objective correlation on the design of benchmark problems, 
in particular when the number of objectives is higher than $2$.
In future works, we will use some sample technics to study the $\rho MNK$-landscapes of larger size.
We will also compare our results on the properties of search space with the performance of different metaheuristics.
However, the efficient set does not cover all the search space properties,
so next works will focus on the properties related to the Pareto local optima, and to the Pareto local optimum sets.

\bibliographystyle{splncs}

\begin{thebibliography}{}

\end{thebibliography}


\begin{thebibliography}{10}

\bibitem{ehrgott1997}
Ehrgott, M., Klamroth, K.:
\newblock Connectedness of efficient solutions in multiple criteria
  combinatorial optimization.
\newblock European Journal of Operational Research \textbf{97}(1) (1997)
  159--166

\bibitem{Mote1991}
Mote, J., Olson, I.M.D.L.:
\newblock A parametric approach to solving bicriterion shortest path problems.
\newblock European Journal of Operational Research \textbf{53}(1) (1991)
  81--92

\bibitem{paquete2006}
Paquete, L., St\"{u}tzle, T.:
\newblock A study of stochastic local search algorithms for the biobjective
  {QAP} with correlated flow matrices.
\newblock European Journal of Operational Research \textbf{169}(3) (2006)
  943--959

\bibitem{knowles2003}
Knowles, J., Corne, D.:
\newblock Instance generators and test suites for the multiobjective quadratic
  assignment problem.
\newblock In: Second International Conference on Evolutionary Multi-Criterion
  Optimization (EMO 2003), Faro, Portugal, Springer. Lecture Notes in Computer
  Science. Volume 2632 (2003)  295--310

\bibitem{aguirre2007}
Aguirre, H.E., Tanaka, K.:
\newblock Working principles, behavior, and performance of {MOEAs} on
  {MNK}-landscapes.
\newblock European Journal of Operational Research \textbf{181}(3) (2007)
  1670--1690

\bibitem{ehrgott2005}
Ehrgott, M.:
\newblock Multicriteria optimization. Second edn.
\newblock Springer (2005)

\bibitem{paquete2007}
Paquete, L., St\"utzle, T.:
\newblock Stochastic local search algorithms for multiobjective combinatorial
  optimization: A review.
\newblock In: Handbook of Approximation Algorithms and Metaheuristics.
  Volume~13 of Computer \& Information Science Series.
\newblock Chapman \& Hall / CRC (2007)

\bibitem{knowles2004}
Knowles, J., Corne, D.:
\newblock Bounded {P}areto archiving: Theory and practice.
\newblock In: Metaheuristics for Multiobjective Optimisation. Volume 535 of
  Lecture Notes in Economics and Mathematical Systems.
\newblock Springer-Verlag (2004)  39--64

\bibitem{gorski2006}
Gorski, J., Klamroth, K., Ruzika, S.:
\newblock Connectedness of efficient solutions in multiple objective
  combinatorial optimization.
\newblock Technical Report 102/2006, University of Kaiserslautern, Department
  of Mathematics (2006)

\bibitem{paquete2009}
Paquete, L., St\"utzle, T.:
\newblock Clusters of non-dominated solutions in multiobjective combinatorial
  optimization: An experimental analysis.
\newblock In: Multiobjective Programming and Goal Programming. Volume 618 of
  Lecture Notes in Economics and Mathematical Systems.
\newblock Springer (2009)  69--77

\bibitem{kauffman93}
Kauffman, S.A.:
\newblock The Origins of Order.
\newblock Oxford University Press, New York (1993)

\bibitem{Hotelling36}
Hotelling, H., Pabst, M.R.:
\newblock Rank correlation and tests of significance involving no assumptions
  of normality.
\newblock Ann. Math. Stat. \textbf{7} (1936)  29--43

\bibitem{WEI:90}
Weinberger, E.D.:
\newblock Correlated and uncorrelatated fitness landscapes and how to tell the
  difference.
\newblock In: Biological Cybernetics. (1990)  63:325--336

\end{thebibliography}

\end{document}